%% file: main.tex
\pdfoutput=1
\documentclass[11pt]{article}
\usepackage[]{acl}
\usepackage{times}
\usepackage{latexsym}

\usepackage{graphicx}
\usepackage{amsmath}
\usepackage{amsfonts}
\usepackage{microtype}
\usepackage{booktabs} 
\usepackage{xspace}
\usepackage{subcaption}
\usepackage{bbm}
\usepackage{color, colortbl}
\usepackage{multirow}
\usepackage{enumitem}
\usepackage{url}
\usepackage[htt]{hyphenat}

\usepackage[T1]{fontenc}
\usepackage[utf8]{inputenc}
\usepackage{microtype}

\def\*#1{\mathbf{#1}}
\definecolor{COLOR_ZS}{HTML}{E6ECE3}

\makeatletter
\renewcommand{\paragraph}{%
  \@startsection{paragraph}{4}%
  {\z@}{3.25ex \@plus 1ex \@minus .2ex}{-0.5em}%
  {\normalfont\normalsize\bfseries}%
}
\makeatother

\title{Is Fine-tuning Needed? Pre-trained Language Models Are Near Perfect for Out-of-Domain Detection}

\author{
\begin{tabular}{ccc}
    Rheeya Uppaal$^1$ & Junjie Hu$^{1,2}$ & Yixuan Li$^1$
\end{tabular}
\\
  $^1$Department of Computer Sciences, \\$^2$Department of Biostatistics and Medical Informatics \\
  University of Wisconsin-Madison \\
  \texttt{\{uppaal, jhu, sharonli\}@cs.wisc.edu} \\
  }

\begin{document}
\maketitle

\begin{abstract}
Out-of-distribution (OOD) detection is a critical task for reliable 
predictions over text. Fine-tuning with pre-trained language models has been a \textit{de facto} procedure to derive OOD detectors with respect to in-distribution (ID) data. Despite its common use, the understanding of the role of fine-tuning and its necessity for OOD detection is largely unexplored. In this paper, we raise the question: \emph{is fine-tuning necessary for OOD detection}? We present a study investigating the efficacy of directly leveraging pre-trained language models for OOD detection, without any model fine-tuning on the ID data. We compare the approach with several competitive fine-tuning objectives, and offer new insights under various types of distributional shifts. Extensive evaluations on 8 diverse ID-OOD dataset pairs demonstrate near-perfect OOD detection performance (with 0\% FPR95 in many cases), strongly outperforming its fine-tuned counterparts. We show that using distance-based detection methods, pre-trained language models are near-perfect OOD detectors when the distribution shift involves a domain change. 
Furthermore, we study the effect of fine-tuning on OOD detection and identify how to balance ID accuracy with OOD detection performance.
Our code is publically available\footnote{\url{https://github.com/Uppaal/lm-ood}}.
\end{abstract}

\input{sections/1-Introduction.tex}
\input{sections/2-Preliminaries.tex}
\input{sections/3-Methods.tex}

\input{sections/4-Experimental-Setup.tex}
\input{sections/5.1-Analysis-1.tex}
\input{sections/5.2-Analysis-2.tex}
\input{sections/5.3-Analysis-3.tex}
\input{sections/5.4-Analysis-4.tex}
\input{sections/6-Related-Work.tex}
\input{sections/7-Discussion.tex}
\input{sections/8-EthicalConsideration.tex}
\input{sections/Limitations}

\section*{Acknowledgements}
Li is supported in part by the AFOSR Young Investigator Award under No. FA9550-23-1-0184; UL Research Institutes through the Center for Advancing Safety of Machine Intelligence; Philanthropic Fund from SFF; and faculty research awards from Google, Meta, and Amazon. Hu is supported in part by a gift fund from ProtagoLabs. Any opinions, findings, conclusions, or recommendations
expressed in this material are those of the authors and do not necessarily reflect the views, policies, or endorsements either expressed or implied, of the sponsors. 
We would like to thank Yifei Ming and the anonymous reviewers for helpful comments.

\bibliography{anthology,custom}

\clearpage
\appendix
\input{sections/Appendix.tex}

\end{document}

%% file: sections/1-Introduction.tex
\section{Introduction}
\label{sec:intro}

Despite recent successes, high-performing pre-trained language models are still fragile under distribution shifts, making their applications to the real world challenging~\cite{ribeiro2020beyond}. In most real-world settings, the train and test distributions are often not independent and identically distributed. Furthermore, test distributions are often non-stationary and can change over time. 
The problem of \textit{out-of-distribution} (OOD) detection addresses the identification of anomalous data, enabling the model to abstain from prediction when it is not supposed to. This is especially important for high-risk settings like financial and medical applications,
where unreliable predictions could incur great costs \cite{ulmer2020trust, zhang2021out}.

In literature, a \textit{de facto} procedure is to fine-tune a pre-trained language model on the in-distribution (ID) data\footnote{Note that the ID data is defined \emph{w.r.t.} the downstream dataset of interest, not the pre-training data.}, and then derive the OOD detector based on the adapted model~\cite{zhou2021contrastive, hendrycks2020pretrained, xu2021unsupervised}.
The fine-tuned model is hypothesized to produce embeddings that are customized to the ID data. Thus, prior work focuses on the design of fine-tuning and expects the adapted representations to be more useful for OOD detection. Despite its common use, the understanding of the role of fine-tuning and its necessity for OOD detection is largely lacking in the field.

Motivated by this, we revisit  the common procedure and raise the unexplored question: \emph{is fine-tuning necessary at all, for OOD detection}? 
To answer this question, we introduce a simple and effective procedure for OOD detection, which does not require any model fine-tuning on the ID data. Specifically, we explore distance-based metrics for detection, which measure the relative distances of samples in the representation space of a pre-trained language model. The operating hypothesis is that embeddings of ID samples are closer to each other than the OOD sample embeddings. To the best of our knowledge, we are the first to explore distance-based OOD detection methods \emph{directly on a pre-trained language model}, rather than the fine-tuned models adopted in previous works.

We show that our method based on a pre-trained language model achieves near-perfect performance in detecting out-of-domain shifts, favorably outperforming its fine-tuned counterparts. 
For example, for \texttt{20NewsGroups} (ID) vs. \texttt{RTE} (OOD), OOD detection with the best fine-tuning loss~\cite{khosla2020supervised} yields an FPR95 of 24.8\%, while a pre-trained language model can perfectly detect \texttt{RTE}
as OOD with 0\% FPR95.
For comprehensive evaluations, we experiment on 8 diverse ID-OOD dataset pairs spanning semantic and background shifts, and show that the strong performance of using the pre-trained model holds consistently. To better understand the strong performance, we further show that pre-trained models display strongly separated domain clusters, both qualitatively and quantitatively. The strong separation of domain clusters leads to the efficacy of distance-based OOD detection. 

Even further, we systematically compare different fine-tuning objectives, and interestingly observe that 
the performance of distance-based OOD detection declines over the course of fine-tuning across all objectives, despite the increase in ID classification accuracy. To this end, we provide new insights that early stopping~\cite{yao2007early} can be a promising solution, if one desires a good tradeoff between  OOD detection and ID classification performance.

Our  contributions
can be summarized as follows:
\begin{enumerate}\itemsep-0.2em 
    \item We propose a simple and effective method for zero-shot\footnote{We use the term ``zero-shot'' to refer to a setting where no (ID or OOD) data is used to update the model parameters.}
    OOD detection, leveraging pre-trained language models without fine-tuning on the ID data. 
   Extensive experiments demonstrate its near-perfect performance (with 0\% FPR95 in most cases), favorably outperforming its fine-tuned counterparts. 
    \item We conduct a comprehensive study to understand  fine-tuning objectives and their impact on OOD detection. We offer new insights on their efficacy under various types of distribution shifts.  
    \item We perform qualitative and quantitative analysis on the embedding characteristics, explaining the strong performance of using a pre-trained language model for OOD detection.
\end{enumerate}

%% file: sections/2-Preliminaries.tex
\section{Preliminaries}
\label{sec:preliminaries}

\paragraph{OOD Detection} For a supervised multi-class classification task, the labeled training dataset $\mathcal{D}_\text{in}=\{(\*x_i,y_i)\}_{i=1}^N$ consists of samples from the joint distribution $P_{\mathcal{X}\mathcal{Y}}$, where $\mathcal{X}$ is the input space and $\mathcal{Y}=\{1,\cdots, C\}$ is the label space. Given a test-time sample $\*x'$, OOD detection aims to identify whether $\*x'$ is in-distribution (ID) $P_\text{in}$ or not, where $P_\text{in}$ is the marginal of $P_{\mathcal{X}\mathcal{Y}}$ on $\mathcal{X}$. Formally, we denote the OOD detector as a binary function mapping $G(\*x'):\mathcal{X} \rightarrow \{\text{in},\text{out}\}$.

 \paragraph{Types of Distribution Shifts} \citet{arora2021types} categorize OOD samples by the type of distribution shift they exhibit in NLP problems. According to \citet{ren2019likelihood}, the representations $h(\*x)$ can be decomposed into two independent and disjoint components---\textit{semantic features} and \textit{background features}. Semantic features are discriminative and strongly correlated with labels for prediction, while background features contain population-level statistics and are invariant across labels.  

Based on the type of features in OOD samples, the distribution shift is categorized as \textit{semantic shift} or \textit{background shift}. An example of the semantic shift is the open-set classification problem that encounters novel classes at test time~\cite{scheirer2012toward}, where the semantic of $\*x'$ is outside the support of $\mathcal{Y}$. Background shift is often seen when the domain or style of texts changes  in the input space $\mathcal{X}$ while $\mathcal{Y}$ remains the same~\cite{pavlick2016empirical}. We comprehensively consider both types of shifts later in our experiments in Section~\ref{sec:exp_setup}.

%% file: sections/3-Methods.tex
\section{Methodology}
\label{sec:methods}

In Section~\ref{sec:zero-shot}, we start by introducing OOD detection with pre-trained language models, which does not require any model fine-tuning on the ID dataset. We further consider OOD detection with model fine-tuning in Section~\ref{sec:fine-tune}.

\subsection{OOD Detection with Pre-trained Models}
\label{sec:zero-shot}

We consider a pre-trained language model backbone $h\colon \mathcal{X} \to \mathbb{R}^{d}$, which encodes an input $\*x$ to a $d$-dimensional text embedding $h(\*x)$. 

The goal of OOD detection is to identify samples that do not belong to $P_\text{in}$. Note that the ID data is defined \emph{w.r.t.} the downstream dataset $\mathcal{D}_\text{in}$ of interest, instead of the pre-training data. 
Different from prior works, \emph{there is no fine-tuning/training on the ID samples}, and the setup is thus labelled as zero-shot OOD detection.

We formulate the zero-shot OOD detector as a binary function mapping:
\begin{equation} \label{eq:G_lambda}
    G_\lambda(\*x;h) = 
\begin{cases}
    \text{in} &\text{if } S(\*x;h) \geq \lambda\\
    \text{out} &\text{if } S(\*x;h) < \lambda\\
\end{cases},
\end{equation} 
where $S(\*x;h)$ is the OOD scoring function, and $\lambda$ is the threshold. By convention, $\lambda$ is chosen so that a high fraction of ID data (\emph{e.g.,} 95\%) is above the threshold. We describe $S(\*x;h)$ in details next. 

We employ distance-based methods for zero-shot OOD detection, which measure the relative distances of samples in representation space. To the best of our knowledge, we are the first to use distance-based OOD detection \emph{directly with a pre-trained language model}, while previous works use models adapted to the ID data. The operating hypothesis is that the embeddings of ID samples are closer to each other than the OOD sample embeddings. Modeling the learned representation space as a mixture of multivariate Gaussians, \citet{lee2018simple} used the {Maximum Mahalanobis distance} \cite{mahalanobis1936generalized} to all class centroids as the score for OOD detection:
\begin{align*}
  S_\text{Maha}(\*x;h) = \min_{c\in \mathcal{Y}} \left( h(\*x) - \boldsymbol\mu_c \right)^\top \\ 
  \Sigma^{-1} \left( h(\*x) - \boldsymbol\mu_c \right), 
\end{align*}
where $\Sigma$ is the covariance matrix and $\boldsymbol\mu_c$ is the mean embedding of class $c$. Both $\Sigma$ and $\boldsymbol\mu_c$ are estimated on the ID embeddings extracted from the pre-trained language model $h(\cdot)$. 

Using Mahalanobis distance for OOD detection requires some distributional assumptions on the representation space. This is circumvented through \emph{non-parametric} density estimation using {nearest neighbors} \cite{sun2022out}. The distance between a query point and its $k$-{th} nearest neighbor in the 
ID
data is used for OOD detection:
\begin{align*}
    S_\text{kNN}(\*x, h) = -\|\*z - \*z_k \|_2,
\end{align*}
where $\*z$ and $\*z_k$ are the $L_2$ normalized embeddings, for the query point $\*x$ and its $k$-{th} nearest neighbor. In Section~\ref{sec:analysis}, we evaluate zero-shot OOD detection performance using both parametric (Maha) and non-parametric (KNN) distance functions. 

\subsection{OOD Detection with Fine-tuning}
\label{sec:fine-tune}

In contrast to the zero-shot OOD detection setup,
an alternative strategy is to fine-tune the model on the ID dataset $\mathcal{D}_\text{in}$ and then perform OOD detection \emph{w.r.t.} the fine-tuned model. In what follows, we comprehensively consider three different fine-tuning objectives: (1) cross-entropy loss, (2) task-adaptive pretraining loss, and (3) supervised contrastive loss.

\paragraph{Cross-Entropy (CE)} 
The cross-entropy loss is widely used for training neural networks, making it an ideal baseline for our study. Given a pre-trained model, we fine-tune with the CE loss:
\begin{align*}
    \mathcal{L}_\text{CE} = \frac{1}{N} \sum_{i=1}^N -\log \frac{e^{f_y(\*x_i;\theta)}}{\sum_{j=1}^C e^{f_j(\*x_i;\theta)}}
\end{align*}
where $f_y$ is the logit output corresponding to the ground truth label $y$, and $\theta$ is the parameterization of the neural network.

\paragraph{Task-adaptive Pretraining (TAPT)} 
\citet{gururangan2020don} show that 
multi-phase adaptive pretraining boosts downstream task performance of pre-trained language models. They introduce Task Adaptive Pre-Training (TAPT), which involves
extending the unsupervised pre-training process (using the masked language modeling objective \cite{devlin2018bert}) with data for the downstream task, before fine-tuning to the same task using cross-entropy. 
TAPT improves generalization capabilities by providing a strong initialization for fine-tuning, and to the best of our knowledge, TAPT has \emph{not} been used in the setting of OOD detection prior to our work.

\paragraph{Supervised Contrastive Learning (SupCon)} By leveraging information on labels and increasing the number of positive pairs during contrastive training, SupCon \cite{khosla2020supervised} has been shown to consistently outperform cross-entropy on large-scale classification tasks \cite{gunel2020supervised}. 
The objective encourages embeddings of a class to be highly separated from other classes, boosting the performance of OOD detection on text classification tasks \cite{zhou2021contrastive}. 
Formally,
\vspace{-0.1cm}
\begin{align*}
\begin{split}
    \mathcal{L}_{\text{SupCon}} &= - \sum_{i=1}^N \frac{1}{N|P(i)|} \\ &\sum_{p \in P(i)} \log \frac{\exp(\*z_i^\top  \*z_p/ \tau)}{\sum_{a \in A(i)} \exp{(\*z_i^\top \*z_a / \tau)}},     
\end{split}
\end{align*}
\noindent where $P(i)$ is the set of anchor instances from the same class as $\*x_i$, $A(i)$ is the set of all anchor instances, $\*z_i$ is the $L_2$ normalized sentence embedding for $\*x_i$, and $\tau$ is the temperature.

After fine-tuning, OOD detection is performed using a similar procedure as Equation~\ref{eq:G_lambda}, except that the scoring function $S(\*x;h)$ is calculated using the fine-tuned model.
While our primary focus is distance-based detection, we additionally consider two common output-based methods---{maximum softmax probability} (MSP) \cite{hendrycks2016baseline} and {energy score} \cite{liu2020energy}. They derive OOD scores from the confidence or logits from the classification head of the model.

%% file: sections/4-Experimental-Setup.tex
\section{Experimental Setup}
\label{sec:exp_setup}

\paragraph{Datasets} We adopt the benchmark in \citet{hendrycks2020pretrained} and \citet{zhou2021contrastive}, examining
9 diverse ID-OOD dataset pairs. Specifically, we use the \texttt{IMDB} dataset~\cite{maas2011learning} and \texttt{SST-2}~\cite{socher2013recursive} on sentiment analysis, the \texttt{20NewsGroups (20NG)} dataset~\cite{lang1995newsweeder} on topic classification, the \texttt{RTE}~\cite{wang2018glue} and \texttt{MNLI}~\cite{williams2018broad} on natural language inference, the English side of \texttt{Multi30k}~\cite{elliott2016multi30k} on machine translation, the cross-intent dataset \texttt{CLINC150}~\cite{larson2019evaluation}, and the \texttt{NewsCategory} multiclass classification dataset~\cite{misra2018news}. Details of the data preparation are described in Appendix \ref{sec:appendix-dataset-setup}.

With these datasets, we examine two main settings: \textit{out-of-domain (OoD) shift} where ID and OOD examples come from different datasets (\emph{i.e.}, domains), and \textit{same-domain (SD) shift} where ID and OOD examples come from the same domain but have disjoint sets of classes. In the OoD setting, we further categorize the ID-OOD pairs into the semantic shift and background shift. Particularly, \texttt{IMDB} and \texttt{SST-2} are both sentiment analysis datasets that have the same set of classes but consist of examples from different domains. In the same-domain setting, we split the \texttt{NewsCategory} dataset, where we make disjoint sets of classes as ID and OOD (Appendix \ref{sec:appendix-dataset-setup}).

\begin{table}[t]
\resizebox{\columnwidth}{!}{%
\begin{tabular}{lll}
\toprule
Settings              & ID             & OOD                          \\ \midrule
\multirow{2}{*}{\textbf{OoD: Semantic Shift}} & \multirow{2}{*}{20NewsGroups} & SST-2, MNLI, RTE, Multi30K \\
                      &                & IMDB, NewsCategory, CLINC150 \\
\textbf{OoD: Background Shift} & IMDB           & SST-2                        \\
\midrule
\textbf{Same Domain Shift}    & NewsCategory-ID & NewsCategory-OOD       \\ \bottomrule       
\end{tabular}}
\caption{Settings of ID-OOD dataset pairs}
\label{tab:settings}
\vskip -0.15in
\end{table}

\input{tables/comparison-of-pretrained-and-ft-models.tex}

\noindent\paragraph{Models} We use RoBERTa~\cite{liu2019roberta}, which is a commonly used pre-trained language model like BERT~\cite{devlin2018bert}. Both models have been used in prior work on OOD detection~\cite{podolskiy2021revisiting, hendrycks2020pretrained}, but we choose RoBERTa as the diverse data it is pre-trained on has been shown to make it stronger for OOD detection~\cite{zhou2021contrastive, podolskiy2021revisiting, hendrycks2020pretrained}. 
{We use embeddings of the beginning-of-sentence (BOS) token as the sentence representation, and compare this to alternate approaches in Appendix \ref{sec:appendix-sent-embeddings}.}
Following ~\citet{zhou2021contrastive}, we fine-tune RoBERTa-base on downstream datasets for 10 epochs. For SupCon, we use a joint objective with Cross Entropy, with weight $\alpha=2$ to the SupCon loss.
For TAPT, we pre-train the model for 3 epochs on the ID data.
For distance-based OOD detection, we use sentence embeddings from the penultimate layer. 
We fine-tune all layers using Adam, with batch size 4, learning rate $10^{-5}$, and weight decay $0.01$. Further details of implementation and configurations are in Appendix~\ref{sec:appendix-checklist-implementation}.

\vspace{-0.4cm}
\noindent\paragraph{Evaluation Metrics} 
We report the following standard metrics: (1) the false positive
rate (\text{FPR}95) of OOD samples when the true positive rate of ID samples is at 95\%, (2) the area under the receiver operating characteristic curve (AUROC), (3) the area under the precision-recall curve (AUPR), and (4) ID classification accuracy (ID ACC). 

%% file: tables/comparison-of-pretrained-and-ft-models.tex
\begin{table*}[ht!] 
\centering
\resizebox{\textwidth}{!}{
\begin{tabular}{llllllllll} 
\toprule
~ & ~ & \multicolumn{4}{c}{\textbf{KNN }(non-parametric)} & \multicolumn{4}{c}{\textbf{Mahalanobis}    
 (parametric)} \\ 

\textbf{ID$\to$OOD Pair} & \textbf{Training} & \textbf{AUROC ↑} & \textbf{AUPR (In) ↑} & \textbf{AUPR (Out) ↑} & \textbf{FPR95 ↓} & \textbf{AUROC ↑} & \textbf{AUPR (In) ↑} & \textbf{AUPR (Out) ↑} & \textbf{FPR95 ↓} \\ \midrule
\multicolumn{10}{l}{\textit{Out-of-Domain: Semantic Shift}} \\ \midrule
        ~ & Zhou et al. & 0.935	& 0.982	& 0.664	& 0.713 & 0.978&	0.994	&0.865&	0.015\\ 
         & CE & 0.973 & 0.991 & 0.923 & 0.155 & 0.981 & 0.994 & 0.942 & 0.087 \\ 
        20NG$\to$SST-2 & TAPT & 0.969 & 0.990 & 0.903 & 0.169 & 0.981 & 0.994 & 0.939 & 0.088 \\
        ~ & SupCon & 0.969 & 0.990 & 0.909 & 0.180 & 0.980 & 0.994 & 0.943 & 0.094 \\ 
        \rowcolor{COLOR_ZS} \cellcolor{white} ~ & Pre-trained & 1.000 & 1.000 & 1.000 & 0.000 & 1.000 & 1.000 & 1.000 & 0.000 \\ \midrule

        ~ & Zhou et al. & 0.935 &	0.929	& 0.950	& 0.718 & 0.964&	0.955	&0.978	&0.224\\ 
         & CE & 0.954 & 0.898 & 0.984 & 0.263 & 0.968 & 0.925 & 0.989 & 0.166 \\ 
        20NG$\to$MNLI & TAPT & 0.950 & 0.887 & 0.982 & 0.263 & 0.964 & 0.910 & 0.988 & 0.175 \\ 
        ~ & SupCon & 0.954 & 0.899 & 0.984 & 0.265 & 0.970 & 0.932 & 0.990 & 0.156 \\ 
        \rowcolor{COLOR_ZS} \cellcolor{white} ~ & Pre-trained  & 1.000 & 0.999 & 1.000 & 0.000 & 1.000 & 0.999 & 1.000 & 0.000 \\ \midrule

        ~ & Zhou et al. & 0.934	& 0.972	& 0.780	& 0.594 & 0.956	&0.981&	0.860&	0.312\\ 
         & CE & 0.922 & 0.958 & 0.858 & 0.410 & 0.945 & 0.970 & 0.902 & 0.285 \\
        20NG$\to$RTE & TAPT & 0.898 & 0.942 & 0.822 & 0.455 & 0.919 & 0.952 & 0.869 & 0.352 \\ 
        ~ & SupCon & 0.923 & 0.959 & 0.858 & 0.393 & 0.952 & 0.975 & 0.914 & 0.248 \\ 
        \rowcolor{COLOR_ZS} \cellcolor{white} ~ & Pre-trained  & 1.000 & 1.000 & 0.999 & 0.000 & 1.000 & 1.000 & 0.999 & 0.000 \\ \midrule

        ~ & Zhou et al. & 0.954	& 0.823	& 0.993	& 0.261 & 0.969	&0.867&	0.996	&0.144 \\ 
         & CE & 0.951 & 0.804 & 0.993 & 0.292 & 0.961 & 0.817 & 0.995 & 0.206 \\ 
        20NG$\to$IMDB & TAPT & 0.955 & 0.797 & 0.994 & 0.227 & 0.965 & 0.804 & 0.995 & 0.159 \\ 
        ~ & SupCon & 0.958 & 0.826 & 0.994 & 0.234 & 0.970 & 0.852 & 0.996 & 0.150 \\ 
        \rowcolor{COLOR_ZS} \cellcolor{white} ~ & Pre-trained  & 0.988 & 0.970 & 0.998 & 0.019 & 0.990 & 0.975 & 0.998 & 0.012 \\ \midrule

        ~ & Zhou et al. & 0.932 & 0.977	& 0.708	& 0.851 & 0.980 &	0.993	& 0.888	& 0.005 \\ 
         & CE & 0.949 & 0.976 & 0.898 & 0.264 & 0.962 & 0.982 & 0.920 & 0.175 \\ 
        20NG$\to$Multi30K & TAPT & 0.940 & 0.970 & 0.886 & 0.258 & 0.956 & 0.978 & 0.922 & 0.167 \\
        ~ & SupCon & 0.937 & 0.969 & 0.887 & 0.294 & 0.955 & 0.977 & 0.918 & 0.201 \\ 
        \rowcolor{COLOR_ZS} \cellcolor{white} ~ & Pre-trained  & 1.000 & 1.000 & 1.000 & 0.000 & 1.000 & 1.000 & 1.000 & 0.000 \\ \midrule

        ~ & Zhou et al. & 0.928 & 	0.921 & 0.937 &	0.765 & 0.955&	0.948	 & 0.969 &	0.383  \\ 
         & CE & 0.939 & 0.877 & 0.977 & 0.339 & 0.957 & 0.905 & 0.984 & 0.234 \\ 
        20NG$\to$NewsCategory & TAPT & 0.931 & 0.853 & 0.973 & 0.343 & 0.947 & 0.874 & 0.981 & 0.243 \\ 
        ~ & SupCon & 0.938 & 0.877 & 0.976 & 0.354 & 0.962 & 0.919 & 0.986 & 0.219 \\ 
        \rowcolor{COLOR_ZS} \cellcolor{white} ~ & Pre-trained  & 1.000 & 0.999 & 1.000 & 0.000 & 1.000 & 0.999 & 1.000 & 0.000 \\ \midrule

        ~ & Zhou et al. & 0.952	& 0.992	& 0.601	& 0.388 & 0.988& 	0.998	& 0.870	& 0.005 \\ 
         & CE & 0.953 & 0.991 & 0.816 & 0.247 & 0.964 & 0.993 & 0.844 & 0.189 \\
        20NG$\to$CLINC150 & TAPT & 0.944 & 0.989 & 0.769 & 0.296 & 0.959 & 0.992 & 0.830 & 0.213 \\
        ~ & SupCon & 0.940 & 0.988 & 0.761 & 0.343 & 0.957 & 0.992 & 0.821 & 0.230 \\ 
        \rowcolor{COLOR_ZS} \cellcolor{white} ~ & Pre-trained  & 1.000 & 1.000 & 1.000 & 0.000 & 1.000 & 1.000 & 1.000 & 0.000 \\
\midrule
\multicolumn{10}{l}{\textit{Out-of-Domain: Background Shift}} \\ \midrule

         & CE & 0.865 & 0.994 & 0.147 & 0.741 & 0.893 & 0.996 & 0.231 & 0.618 \\ 
        IMDB $\to$ SST-2 & TAPT & 0.857 & 0.994 & 0.137 & 0.746 & 0.877 & 0.995 & 0.172 & 0.683 \\ 
        ~ & SupCon & 0.838 & 0.993 & 0.119 & 0.824 & 0.865 & 0.995 & 0.149 & 0.800 \\ 
       \rowcolor{COLOR_ZS} \cellcolor{white} ~ & Pre-trained & 0.967 & 0.999 & 0.582 & 0.210 & 0.996 & 1.000 & 0.860 & 0.004 \\
\midrule
\multicolumn{10}{l}{\textit{Same Domain Shift}} \\ \midrule

         & CE & 0.925 & 0.922 & 0.933 & 0.465 & 0.877 & 0.815 & 0.912 & 0.467 \\
        NewsCategory-ID $\to$ & TAPT & 0.918 & 0.917 & 0.924 & 0.513 & 0.876 & 0.822 & 0.907 & 0.502 \\ 
        NewsCategory-OOD & SupCon& 0.925 & 0.922 & 0.933 & 0.465 & 0.877 & 0.815 & 0.912 & 0.467 \\ 
        \rowcolor{COLOR_ZS} \cellcolor{white} ~ & Pre-trained & 0.816 & 0.839 & 0.806 & 0.845 & 0.550 & 0.458 & 0.628 & 0.939 \\
       
\bottomrule
\end{tabular}}
\caption{Comparison of OOD detection performance of pre-trained and fine-tuned models. Pre-trained language models are near-perfect OOD detectors in the out-of-domain setting, but worst in the same-domain shift setting.}
\label{tab:comparison-of-pretrained-and-ft-models}
\end{table*}

\begin{figure*}[ht]
\begin{center}
\centerline{
\includegraphics[width=0.5\columnwidth]{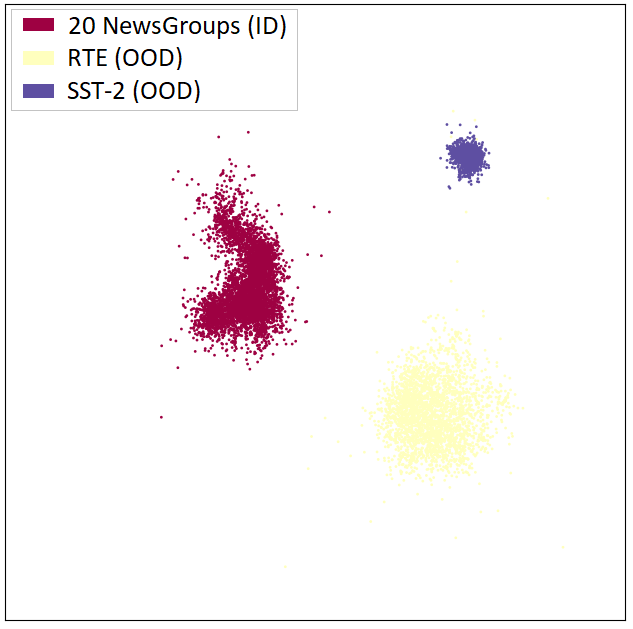}
\includegraphics[width=0.5\columnwidth]{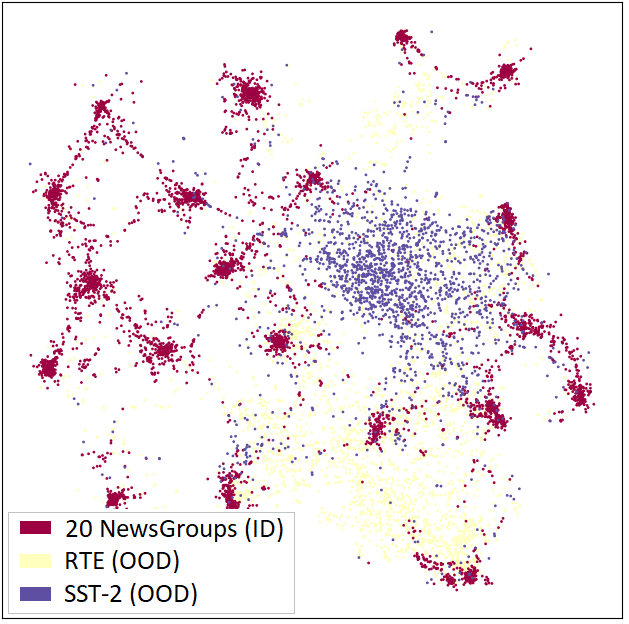}
\includegraphics[width=0.5\columnwidth]{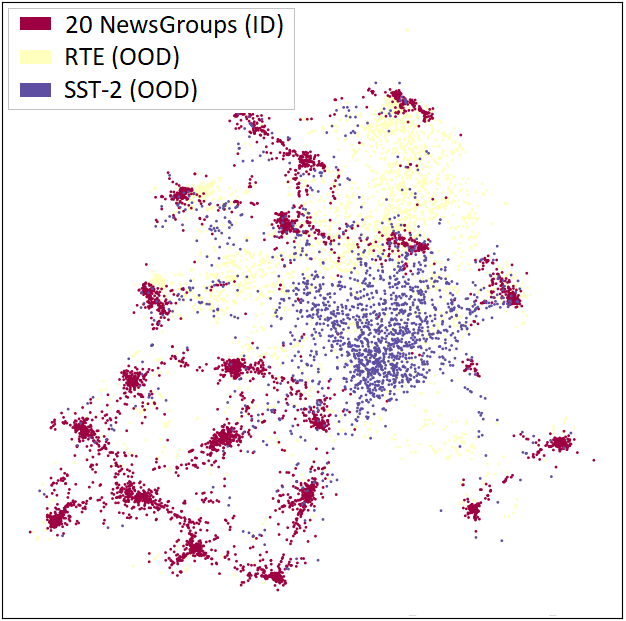}
\includegraphics[width=0.5\columnwidth]{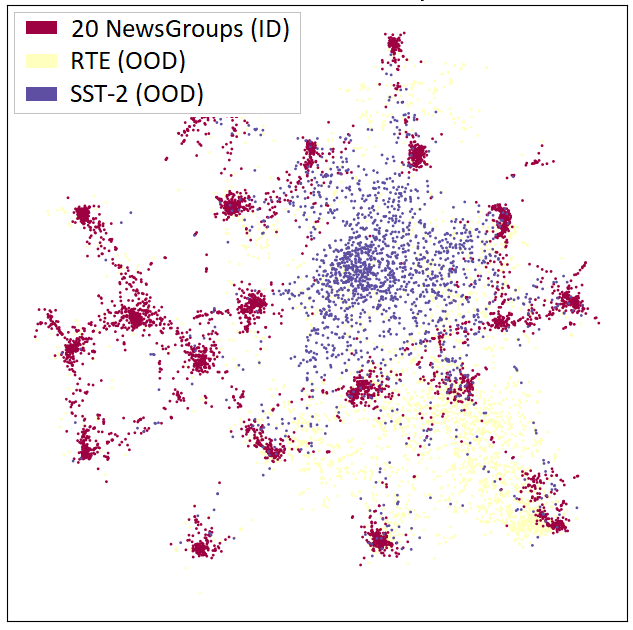}
}
\caption{Comparison of data representations from the penultimate layer of pre-trained and fine-tuned models. \textbf{From left to right}: (1) Pre-trained model, (2) Fine-tuning with Cross-Entropy (CE), (3) Fine-tuning with TAPT, and (4) Fine-tuning with SupCon. The ID dataset, 20NewsGroups, is shown in {\color{purple}maroon}, while the OOD datasets RTE and SST-2 are in {\color{yellow}yellow} and {\color{violet}purple} respectively. The pretrained model represents each domain as a separate cluster, strengthening distance-based OOD performance. Fine-tuning encourages the model to learn class-specific clusters, making distance based OOD detection more challenging.}
\label{fig:comparison-of-pretrained-and-ft-models-20ng}
\end{center}
\end{figure*}

%% file: sections/5.1-Analysis-1.tex
\section{Results and Analysis}
\label{sec:analysis}

\subsection{Out-of-domain detection with pre-trained language models is near perfect}
\label{sec:analysis-is-ft-important}

Table \ref{tab:comparison-of-pretrained-and-ft-models}
shows the pre-trained model outperforming all its fine-tuned variants in the out-of-domain shift setting, and achieving near-perfect OOD detection on all ID-OOD pairs considered. {In addition to comparisons with three fine-tuning objectives, we also compare with a competitive baseline proposed by \citet{zhou2021contrastive}, which fine-tunes a model with a novel contrastive objective.}
Taking \texttt{20NewsGroups} (ID) vs. \texttt{RTE} (OOD) as an example, OOD detection with the best fine-tuning strategy (\emph{i.e.}, SupCon) yields an FPR95 of 24.8\%. In sharp contrast, zero-shot OOD detection using the pre-trained language model can perfectly detect RTE as OOD with \textbf{0\% FPR95}. We investigate same-domain shift in-depth later in Section~\ref{sec:same-domain}. 

Figure \ref{fig:comparison-of-pretrained-and-ft-models-20ng} sheds some light on the strong performance of pre-trained language models for out-of-domain detection. In the leftmost figure, we observe that large pre-trained language models create separate domain clusters of sentence embeddings for ID and OOD data, matching the findings of \citet{aharoni2020unsupervised}. The strong separation of clusters boosts the performance of distance-based OOD detection. In contrast, fine-tuning induces a model to divide a single domain cluster into multiple class clusters. When a fine-tuned model encounters an OOD datapoint, it attempts to classify it by mapping it to one of the existing ID class clusters. However, due to the distributional difference of the datapoint, the model is unable to perfectly map such a point and OOD points end up in the space between the ID class clusters most similar to it.
Fine-tuned representations of the data thus make distance-based OOD detection more challenging.

%% file: sections/5.2-Analysis-2.tex

\begin{figure*}[ht]
\vskip 0.2in
\begin{center}
\centerline{\includegraphics[width=2\columnwidth]{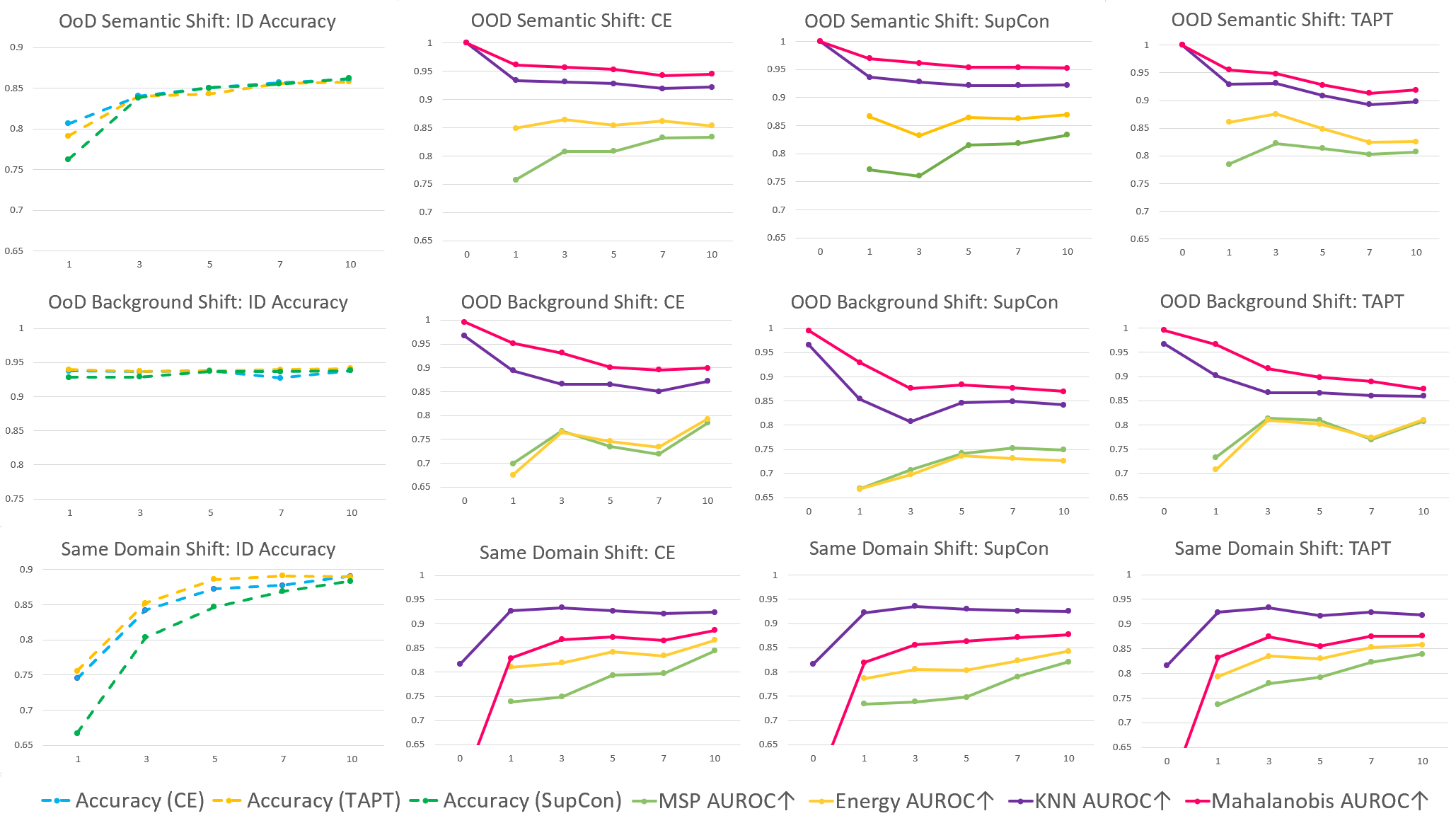}}
\caption{Effect of fine-tuning on ID accuracy and OOD detection performance, across different objectives and detection methods. From left to right: (1) ID Accuracy, AUROC with (2) CE, (2) TAPT, and (3) SupCon losses. From top to bottom: OoD semantic shift, OoD background shift, and same-domain (SD) shift. The X-axis shows the number of fine-tuning epochs, with `0' indicating the pre-trained model. The Y-axis shows either the ID accuracy or the AUROC. Actual values can be found in Appendix~\ref{sec:appendix-finetuning-trajectories}.}

\label{fig:finetuning-trajectories}
\end{center}
\vskip -0.3in
\end{figure*}

\subsection{What's the best way of fine-tuning for OOD detection?}
\label{sec:fine-tuning-analysis}

While pre-trained models show strong out-of-domain detection performance, they lack the classification ability on the ID dataset. This is expected since the models are not optimized for the downstream classification task. Thus, we raise the next question: \emph{How can we fine-tune the model to accurately classify ID data while having reasonable OOD detection performance?} 

To answer this question, we comprehensively compare three fine-tuning objectives (\emph{c.f.} Section~\ref{sec:fine-tune}), coupled with different OOD detection methods. Figure \ref{fig:finetuning-trajectories} depicts the effect of fine-tuning for OOD detection, for both semantic shift (top: \texttt{20NewsGroups} vs. \texttt{RTE}) and background shift (middle: \texttt{IMDB} vs. \texttt{SST-2}).   We highlight three key observations:
\textbf{(1)} For distance-based methods, the OOD detection performance worsens as the number of fine-tuning epochs increases, highlighting that early stopping is the key to strong OOD detection performance. For example, on \texttt{20NewsGroups} (ID) vs. \texttt{RTE} (OOD), the model trained with TAPT for 1 epoch yields an AUROC of 95.5\% (with Mahalanobis), which declines to 91.9\% after 10 epochs of fine-tuning. To the best of our knowledge, we are the first to show the importance of early stopping on fine-tuning language models for distance-based OOD detection. \textbf{(2)} Irrespective of the fine-tuning objectives, distance-based OOD detection methods consistently outperform output-based methods, particularly MSP using softmax confidence~\cite{hendrycks2016baseline} and energy score using logits~\cite{liu2020energy}. \textbf{(3)} Under semantic shift,  out-of-domain detection using any of the three fine-tuning objectives displays similar performance on most ID-OOD pairs, bearing a large gap \emph{w.r.t.} the pre-trained language model. 

\vspace{-0.3cm}
\paragraph{Linear Probing is Suboptimal} 
To perform classification while preserving the OOD detection performance of a pre-trained model, one possible solution is linear probing~\cite{alain2017understanding}, \emph{i.e.,} fine-tuning the classification head to the downstream task, while keeping the weights of the pre-trained model backbone unchanged. However, in Figure~\ref{fig:20ng-linear-probing-acc} (Appendix), we show that linear probing does not yield competitive classification performance. In particular, we observe the strongest fine-tuning objective (TAPT) only obtains an ID accuracy of  61\% after 100 epochs of fine-tuning, compared to full network fine-tuning where an accuracy of 86\% is achieved in 10 epochs. 

%% file: sections/5.3-Analysis-3.tex

\subsection{Investigation on same-domain data shifts}
\label{sec:same-domain}

In this subsection, we further investigate a more challenging type of data shift, where the test samples are from the \emph{same domain} and thus can be distributionally very close to the ID data. This is in contrast to our evaluations in Sections~\ref{sec:analysis-is-ft-important} and~\ref{sec:fine-tuning-analysis}, where the OOD samples are from different domains. To simulate same-domain shifts, we split the \texttt{NewsCategory} dataset into two sets with disjoint classes: one for ID, and another for OOD. The domain for both sets of classes is identical, while the semantic label sets are different. The allocation of classes is described in Table \ref{tab:nc-class-allocation} (Appendix \ref{sec:appendix-dataset-setup}).

\begin{figure}[ht]
\begin{center}
\centerline{
\includegraphics[width=0.5\columnwidth]{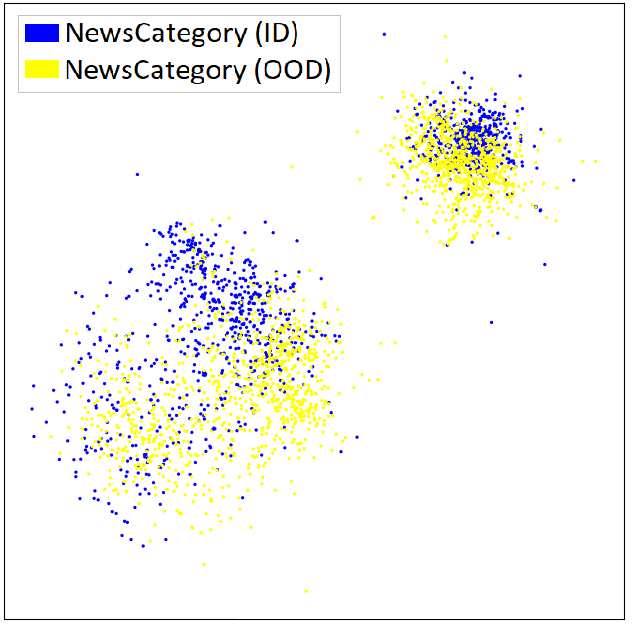}
\includegraphics[width=0.5\columnwidth]{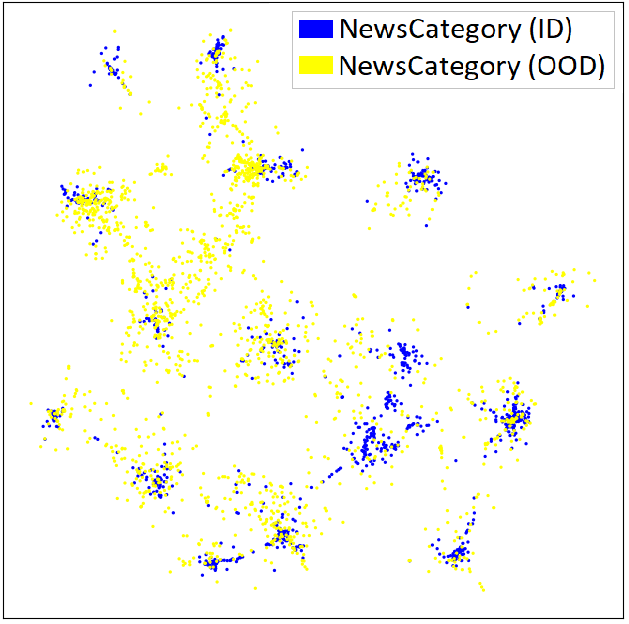}
}
\caption{Comparison of data representations in the penultimate layer of pre-trained vs. fine-tuned models for \emph{same-domain} data shifts. Here we split the \texttt{NewsCategory} dataset into two parts with disjoint classes: one for ID, and another for OOD. ID data is shown in {\color{blue}blue}, while OOD data is in {\color{yellow}yellow}. \textbf{Left}: Pre-trained model. \textbf{Right}: Fine-tuned with cross-entropy loss. Fine-tuning encourages the model to separate the embeddings into individual class clusters.}
\label{fig:comparison-of-pretrained-and-ft-models-nc}
\end{center}
\vskip -0.2in
\end{figure}

Figure \ref{fig:finetuning-trajectories} (bottom) shows the effect of fine-tuning for detection in this challenging setup of same-domain shifts. A salient observation is that fine-tuning consistently improves OOD detection performance, across all training objectives. To better understand why the pre-trained model underperforms in this case, in Figure~\ref{fig:comparison-of-pretrained-and-ft-models-nc}, we plot feature representations, before and after fine-tuning, respectively. As seen in the left of Figure \ref{fig:comparison-of-pretrained-and-ft-models-nc}, when both ID and OOD data are sampled from the same domain, their embeddings are highly overlapping. This explains the suboptimal performance of directly employing embeddings from the pre-trained language model. In contrast, fine-tuning creates stronger separability between ID and OOD data. Table~\ref{tab:sd-seperability}
quantitatively confirms that fine-tuning leads to stronger ID-OOD separability (\emph{c.f.} Equation~\ref{eq:separability}).

\begin{table}
\centering
\resizebox{0.4\textwidth}{!}{
\begin{tabular}{ll} 
\toprule
        \textbf{Training} & \textbf{ID-OOD Separability ↑} \\ \midrule
        CE & 12.235 \\ 
        TAPT & {12.489} \\
        SupCon & 7.549 \\
        Pre-trained & 0.138 \\ \bottomrule
\end{tabular}
}
\caption{Effect of fine-tuning on ID-OOD separability, for same-domain (SD) shift with the \texttt{NewsCategory} dataset. Fine-tuning for a single epoch helps separate overlapping ID and OOD data into dispersed clusters.}
\label{tab:sd-seperability}
\vskip -0.1in
\end{table}

%% file: sections/5.4-Analysis-4.tex
\subsection{Deeper look at embedding quality}

We quantitatively measure the embeddings produced by both pre-trained and fine-tuned language models. We adopt the following three metrics as in~\citet{ming2023cider}: (1) inter-class dispersion, which is the average cosine similarity among pair-wise class centroids, (2) intra-class compactness, which measures the average cosine similarity between each feature embedding and its corresponding class centroid, and (3) ID-OOD separability, which functions as a measure of domain gap between ID and OOD. Formally, 
\begin{align*}
  \text{Disp.}
  (\uparrow) = \frac{1}{C} \sum_{i=1}^C \frac{1}{C-1} \sum_{j=1}^C \boldsymbol\mu_i \cdot \boldsymbol\mu_j  \mathbbm{1} \{i\neq j\} \\
  \text{Comp.}
  (\downarrow)= \frac{1}{C} \sum_{j=1}^C \frac{1}{N} \sum_{i=1}^N \*z_i \cdot \boldsymbol\mu_j \mathbbm{1}\{y_i = j\}
\end{align*}
\begin{align}
\begin{split}
\label{eq:separability}
  \text{Sep.}
  (\uparrow) &= \frac{1}{|\mathcal{D}_\text{out}^\text{test}|} \sum_{\*x' \in \mathcal{D}_\text{out}^\text{test}} \max_{j \in \mathcal{Y}} \*z_{\*x'} \cdot \boldsymbol\mu_j \\
  &- \frac{1}{|\mathcal{D}_\text{in}^\text{test}|} \sum_{\*x \in \mathcal{D}_\text{in}^\text{test}} \max_{j \in \mathcal{Y}} \*z_{\*x} \cdot \boldsymbol\mu_j,  
  \end{split}
\end{align}
where $\boldsymbol\mu_i$ is the average of embeddings for samples in class $i$, and $\*z$ is the $L_2$ normalized embedding.

\begin{table}[t]
\centering
\resizebox{0.5\textwidth}{!}{
\begin{tabular}{llllll} 
\toprule
    \multicolumn{2}{c}{\textbf{ID}} & \textbf{Objective} & \textbf{ID Accuracy ↑}& \textbf{Dispersion ↑} & \textbf{Compactness ↓} \\
    & & & & (in degree) & (in degree)\\
    \midrule
         & & CE & 0.791 & 90.994 & 19.575 \\ 
        \textbf{20NewsGroups} & ~ & TAPT & \textbf{0.807} & \textbf{91.753} & 18.902 \\ 
        & ~ & SupCon & 0.763 & 89.354 & 21.987 \\ 
        & ~ & Pre-trained & 0.053 & 1.514 & \textbf{4.326} \\ \midrule
        & & CE & 0.938 & 87.041 & 21.787 \\ 
        \textbf{IMDB} & ~ & TAPT & \textbf{0.940} & 76.871 & 15.894 \\ 
        & ~ & SupCon & 0.928 & \textbf{135.550} & 19.245 \\ 
        & ~ & Pre-trained & 0.500 & 0.636 & \textbf{6.058} \\ \midrule
        & & CE & 0.745 & \textbf{88.701} & 33.878 \\ 
        \textbf{NewsCategory}  & ~ & TAPT & \textbf{0.756} & 88.216 & 33.509 \\ 
        & ~ & SupCon & 0.667 & 63.392 & 30.793 \\
        & ~ & Pre-trained & 0.050 & 3.086 & \textbf{9.210} \\ \bottomrule
\end{tabular}}
\caption{Quality of ID embeddings generated by pre-trained and fine-tuned models, quantified by accuracy on the ID test set, inter-class dispersion, and intra-class compactness. The fine-tuned models show well-separated and compact class clusters, while the pre-trained model shows a single domain cluster, a sub-optimal setting for downstream classification. Fine-tuned models are trained for a single epoch.}
\label{tab:pretrained-vs-finetuned-id-metrics}
\vskip -0.2in
\end{table}

\begin{figure}[t]
\vskip 0.2in
\begin{center}
\centerline{
\includegraphics[width=\columnwidth]{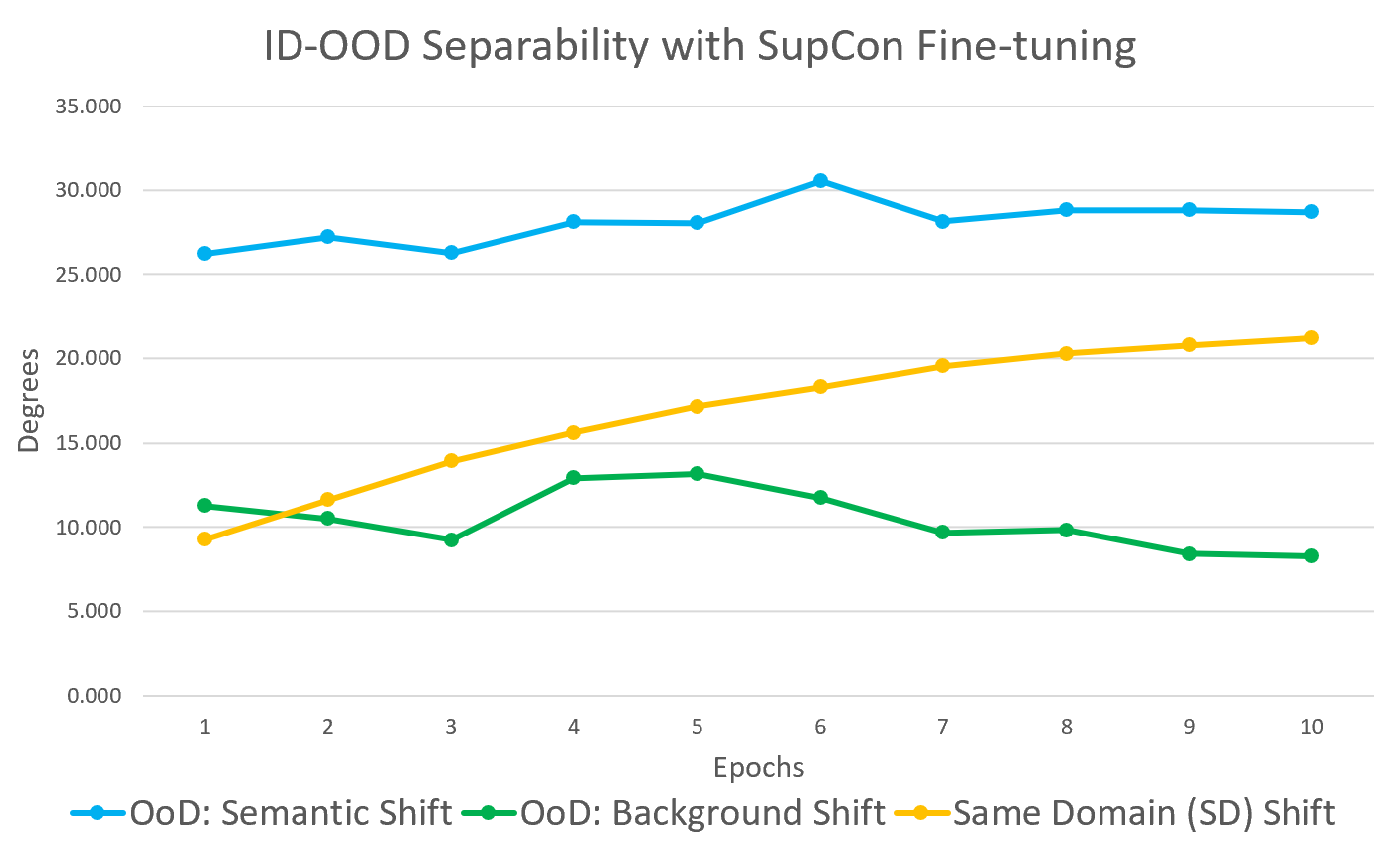}
}
\caption{Effect of fine-tuning (w/ SupCon loss) on the ID-OOD separability. The X-axis shows the number of fine-tuning epochs, and the Y-axis shows ID-OOD separability (in angular degrees). }
\label{fig:embedding-analysis}
\end{center}
\vskip -0.1in
\end{figure}

Table \ref{tab:pretrained-vs-finetuned-id-metrics} shows us that fine-tuning encourages the model to embed the data into well-separated class clusters with high inter-class dispersion (measured in angular degrees). In contrast, the pre-trained model represents the entire domain as a homogeneous cluster containing data from all classes. Interestingly, the pre-trained model displays the strongest compactness, indicating the closeness among ID data points in the original representation space. Note that the ID accuracy is random for the pre-trained model, which is expected.
Dispersion and compactness monotonically improve through fine-tuning, further indicating that fine-tuning encourages the model to project the data into well-separated and compact class-wise clusters. However, Figure \ref{fig:embedding-analysis} shows us that while fine-tuning improves ID-OOD separability for the same-domain shift, it has less impact on out-of-domain shifts. (Actual values and results for other objectives can be found in Appendix~\ref{sec:appendix-finetuning-trajectories}.) This trend also echos our previous observations in Section~\ref{sec:fine-tuning-analysis} and Section~\ref{sec:same-domain}, on OOD detection performance.

%% file: sections/6-Related-Work.tex
\section{Related Work}
\label{sec:related}

The problem of OOD detection  is different from domain adaptation \cite{ramponi2020neural}, where a model is trained to generalize to a known target domain with the same label space. It is also different from
selective prediction where a model abstains only when its confidence is low, irrespective of domain \cite{el2010foundations, geifman2017selective, kamath2020selective}. 

\noindent\paragraph{OOD Detection Methods} A popular baseline is the calibration method Maximum Softmax Probability (MSP)~\cite{hendrycks2016baseline}, 
that directly uses maximum class probability produced by the logits of a trained classifier. However, predictive confidence has been shown to be undesirably high for OOD samples, making MSP ineffective~\cite{nguyen2015deep, wei2022mitigating, shen2021enhancing}.  \citet{liu2020energy} propose using energy score for OOD detection, which better distinguishes in- and out-of-distribution samples than softmax scores.  {ReAct} \cite{sun2021react} improves the energy score by introducing a rectified activation, which reduces model overconfidence in OOD data. \citet{sun2022dice} utilize logit sparsification to enhance the vanilla energy score. 
More recently, detection methods that utilize distances of samples in representation space, have risen as a promising class of OOD detection methods in both the vision \cite{mandelbaum2017distance, lee2018simple, sun2022out, ming2023cider} and multi-modal \cite{mingdelving} regimes.

\noindent\paragraph{OOD Detection in NLP} In the realm of NLP, model confidence using sentence embeddings 
has been shown to be a strong baseline with pre-trained transformers \cite{hendrycks2020pretrained, desai2020calibration}. 
Contrastive learning \cite{khosla2020supervised, gao2021simcse, jin2022towards} minimizes intra-class variance, leading to stronger OOD detection, especially in low data regimes \cite{zeng2021modeling}, and with Mahalanobis distance \cite{zhou2021contrastive, podolskiy2021revisiting}.
Detection performance has also been strengthened using data augmentation \cite{chen2021gold, rawat2021pnpood}, discriminative training \cite{zhan2021out}, mutual information maximization \cite{nimah2021protoinfomax}, ensembles \cite{li2021kfolden} and prototypical networks in the few-shot setup \cite{tan2019out}. While most previous works perform fine-tuning on the ID data, we provide a comprehensive understanding on  \emph{directly using the pre-trained model for zero-shot OOD detection}. 

\vspace{-0.3cm}
\noindent\paragraph{Pre-trained vs Fine-tuned} Pre-trained language models have been shown to learn implicit sentence representations, forming unsupervised domain clusters \cite{aharoni2020unsupervised}.  \citet{andreassen2021evolution} and \citet{kumar2021fine} showed that fine-tuning distorts pre-trained features, worsening accuracy on OOD generalization. However, to the best of our knowledge, we are the first to explore the effect of directly using pre-trained language models for \emph{OOD {detection}}. Related to our work, \citet{mingdelving} show that pre-trained models can be used for zero-shot OOD detection. Different from ours, they perform OOD detection in the multi-modal space and calculate distances between the visual and textual representations. 

%% file: sections/7-Discussion.tex
\section{Conclusion}
\label{sec:discussion}

In this paper, we explore the simple and effective setting of zero-shot OOD detection with pre-trained langage models.
Our work departs from prior literature that typically requires fine-tuning on the ID data. Extensive evaluations demonstrate that pre-trained models are near-perfect for OOD detection when the test data comes from a different domain. We additionally investigate the effect of fine-tuning on OOD detection, and identify strategies to achieve both strong OOD detection performance and ID accuracy. We perform both qualitative and quantitative analysis on the embedding characteristics, explaining the strong performance of our method.
We hope our work will inspire future work to the strong promise of using pre-trained models for OOD detection.

%% file: sections/8-EthicalConsideration.tex
\section*{Ethical Considerations}
\label{sec:appendix-checklist-ethical-considerations}

Our project aims to improve the reliability and safety of large  language models, which can be fragile under distribution shift~\cite{ribeiro2020beyond} and incur great costs \cite{ulmer2020trust, zhang2021out}. By properly flagging anomalous data, our
method can lead to direct benefits and societal impacts, particularly for safety-critical applications. From a user's perspective, our method can help improve trust in the language models. 
Our study does not involve any human subjects or violation of legal
compliance. We do not anticipate any potentially harmful consequences to our work. As detailed in Appendix~\ref{sec:appendix-dataset-setup}, all of our experiments are  conducted using publicly available datasets. Our code has been released for reproducibility. Through our
study and releasing our code, we hope to raise stronger research and societal awareness toward the
problem of out-of-distribution detection in natural language processing.

%% file: sections/Limitations.tex
\section*{Limitations}
\label{sec:appendix-checklist-limitations}

We provide a comprehensive study on the efficacy of leveraging pre-trained language models for zero-shot OOD detection. Our method is thus limited to the setting of abstaining from prediction on all OOD data. This is more conservative than selective prediction, where the model must make predictions over as many ID \& OOD points as possible while maintaining high accuracy. 
Despite this, OOD detection has lower risks to high-risk and safety-critical applications, where rare and anomalous data is more reasonably flagged to the expert.
We believe our work provides new values and insights to the research community, especially on safe handling of distributional shifts when deploying pre-trained language models.

As discussed in our Ethical Considerations, the OOD detection problem is of significant use in high-risk settings, and should be incorporated into production-level pipelines. However, for the same reason, the OOD detection models must be also reliable to avoid any risk to the downstream applications.

%% file: sections/Appendix.tex
\input{sections/Appendix/Preparation-of-Evaluation-Benchmarks}
\input{sections/Appendix/Ablation-on-the-Effect-of-Layers}
\input{sections/Appendix/Sentence-Embeddings}

\input{sections/Appendix/Detailed-Performance-of-Fine-tuning-for-OOD-Detection}

\begin{figure}[ht]
\vskip 0.2in
\begin{center}
\centerline{
\includegraphics[width=\columnwidth]{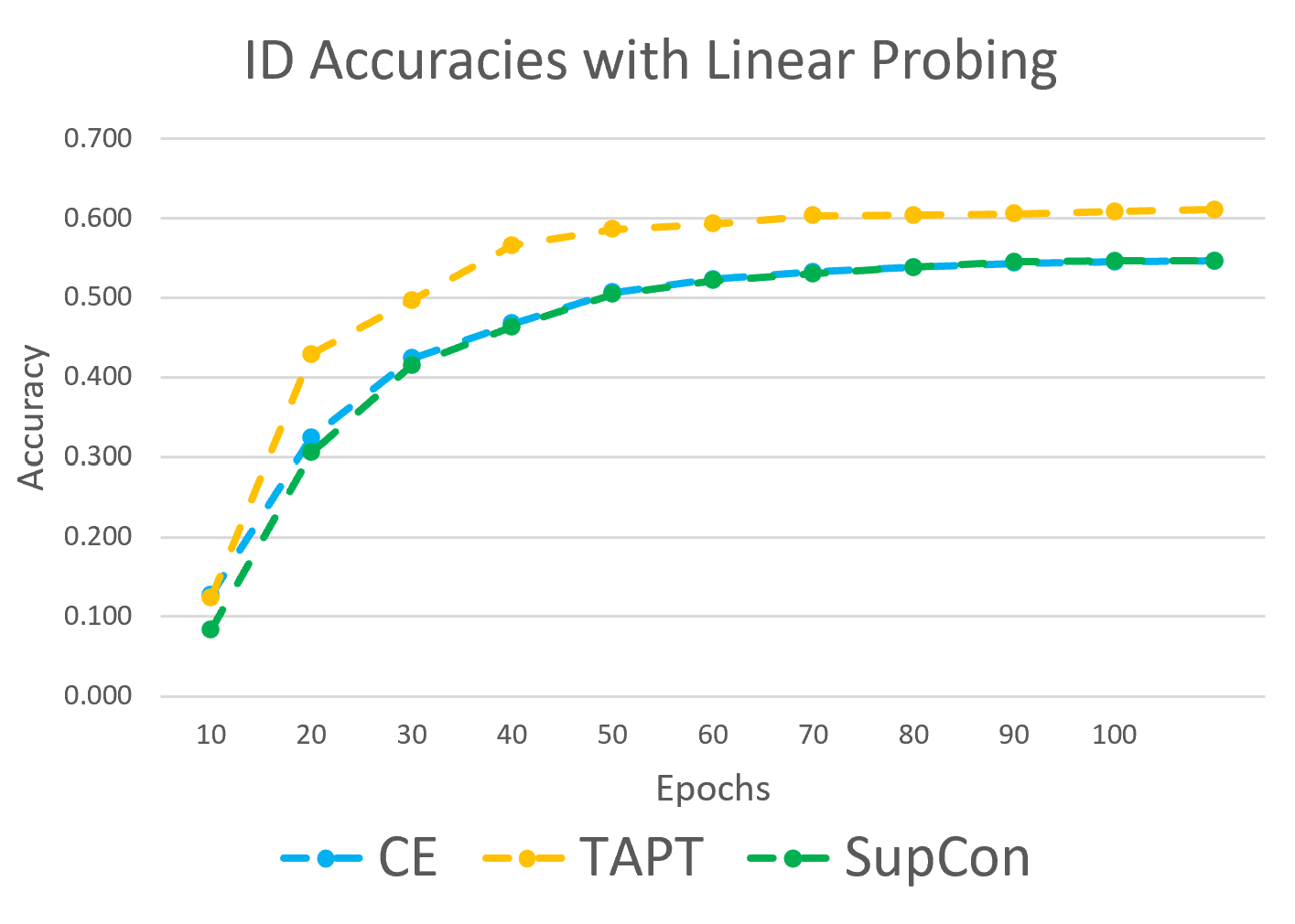}
}
\caption{ID accuracy with linear probing instead of fine-tuning, with \texttt{20NewsGroups}. In comparison to fine-tuning with TAPT, where the accuracy after 10 epochs is 86\%, linear probing with TAPT achieves an accuracy of about only 61\% after 100 epochs.}
\label{fig:20ng-linear-probing-acc}
\end{center}
\vskip -0.35in
\end{figure}

\input{sections/Appendix/Effect-of-Temperature-in-SupCon}
\input{sections/Appendix/Effect-of-k}
\input{tables/implementation-details.tex}
\input{sections/Appendix/Details-on-Implementation}

\input{tables/supcon-temperature-effect}

%% file: sections/Appendix/Preparation-of-Evaluation-Benchmarks.tex
\section{Preparation of Evaluation Benchmarks}
\label{sec:appendix-dataset-setup}

For ID data, we use the train splits of the \texttt{IMDB} dataset on sentiment analysis~\cite{maas2011learning}, and the \texttt{20NewsGroups} dataset on topic classification~\cite{lang1995newsweeder}. For OOD data, we use the test splits of \texttt{IMDB} and \texttt{20NewsGroups}, as well as the test splits from the sentiment classification dataset \texttt{SST-2}~\cite{socher2013recursive}, Natural Language Inference datasets \texttt{RTE}~\cite{wang2018glue} and \texttt{MNLI}~\cite{williams2018broad}, the English source side of machine translation dataset \texttt{Multi30k}~\cite{elliott2016multi30k}, and the cross intent dataset \texttt{CLINC150} \cite{larson2019evaluation}. For \texttt{MNLI}, we use both the matched and mismatched test sets. For \texttt{Multi30k}, we combine the flickr 2016 English test set, mscoco 2017 English test set, and filckr 2018 English test. For \texttt{CLINC150}, we use the `out of scope' class as the test set.

\begin{figure}[ht]
\vskip 0.2in
\begin{center}
\centerline{\includegraphics[width=1.0\columnwidth]{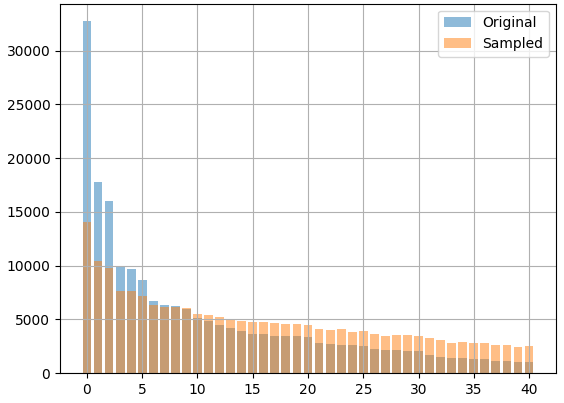}}
\caption{Class frequencies of the \texttt{NewsCategory} dataset. The original frequencies in {\color{blue}blue} show a strong class imbalance, while the modified frequencies in {\color{orange}orange} are more balanced.}
\label{fig:nc-class-frequencies}
\end{center}
\vskip -0.3in
\end{figure}

Inspired by \citet{arora2021types}, we evaluate the detection performance under same-domain shift using the \texttt{NewsCategory}~\cite{misra2018news} dataset. We create two disjoint sets of classes, used as ID and OOD respectively. The domain for both sets of classes is identical, while the label sets differ. Notably, the \texttt{NewsCategory} dataset contains classes with similar semantics, for example `Arts' and `Arts \& Culture'. To ensure the semantic distinction between the ID and OOD classes, we categorize semantically similar classes to be  entirely in either ID or OOD sets. The allocation of classes is summarized in Table \ref{tab:nc-class-allocation}. The dataset also has a strong class imbalance, so we sample data points according to a multinomial distribution, following \citet{lample2019cross}. Figure \ref{fig:nc-class-frequencies} shows the class frequencies before and after sampling. 

More statistics about each dataset is available in Table~\ref{tab:dataset-details}. The listed datasets are intended for research purposes only. We do not make any commercial use of them. 

\begin{table}
\centering
\resizebox{0.3\textwidth}{!}{
    \begin{tabular}{ll}
    \toprule
        ID Classes & OOD Classes \\ \midrule
        Politics & Style \& Beauty \\ 
        The Worldpost & Style \\
        Worldpost & Arts \\ 
        World News & Arts \& Culture \\ 
        Impact & Culture \& Arts \\
        Crime & Food \& Drink \\
        Media & Taste \\ 
        Business & College \\ 
        Money & Education \\ 
        Fifty & Science \\ 
        Good News & Tech \\ 
        Queer Voices & Sports \\ 
        Black Voices & Wellness \\ 
        Women & Healthy Living \\ 
        Latino Voices & Travel \\ 
        Religion & Home \& Living \\ 
        Weird News & Parenting \\ 
        ~ & Parents \\ 
        ~ & Weddings \\ 
        ~ & Divorce \\ 
        ~ & Entertainment \\ 
        ~ & Comedy \\ 
        ~ & Environment \\ 
        ~ & Green \\ 
        \bottomrule
    \end{tabular}
    }
\caption{Division of classes in the \texttt{NewsCategory} dataset into disjoint ID and OOD sets.}
\label{tab:nc-class-allocation}
\end{table}


\begin{table*}[ht]
\centering
\resizebox{\textwidth}{!}{
    \begin{tabular}{lllllll}
    \toprule
        \textbf{Dataset} & \textbf{Domain} & \textbf{Language} & \textbf{License} & \multicolumn{3}{c}{\textbf{Statistics}} \\ 
        ~ & ~ & ~ & ~ & Train & Val & Test \\ \midrule
        \texttt{IMDB} & Large Movie Review Dataset & English & Unknown & 25,000 & 25,000 & 50,000\\
        \texttt{20NewsGroups} & News Articles & English & Unknown & 11314 & 2000 & 5532\\
        \texttt{SST-2} & Movie Reviews & English & cc-by-4.0 & 67349 & 872 & 1821\\
        \texttt{RTE} & News and Wikipedia text & English & cc-by-4.0 & 2490 & 277 & 3000\\
        \texttt{MNLI} & Open American National Corpus & English & cc-by-4.0 & 392702 & 19647 & 19643\\ 
        \texttt{Multi30k} & Flickr30K, MSCOCO & English, German & Custom (research-only, non-commercial) & N/A & N/A & 2532\\
        \texttt{CLINC150} & Intent Classification & English & cc-by-3.0 & 15000 & 3000 & 1000 \\ 
        \texttt{NewsCategory} & HuffPost & English & CC0: Public Domain & 64856 & 4053 & 17968\\    
        \bottomrule
    \end{tabular}
    }
\caption{Artifacts used in our study. The dataset statistics report the values used in our study. For example, the values of the \texttt{NewsCategory} dataset are reported after sampling.}
\label{tab:dataset-details}
\end{table*}

%% file: sections/Appendix/Ablation-on-the-Effect-of-Layers.tex
\section{Ablation on the Effect of Layers}
\label{sec:appendix-which-layer-representation-is-best}

The RoBERTa architecture consists of a backbone of multiple transformer layers, followed by a task-specific head on top. For the classification task, this task-specific head consists of a dense layer followed by a classification projection layer. \citet{zhou2021contrastive} use the features from after the dense layer for OOD detection. Instead, we use the features from before this layer. Table~\ref{tab:comparison-of-ft-models-before-dense}
shows the OOD detection performance using the representations from after the dense layer. Table~\ref{tab:comparison-of-ft-models-before-dense} displays a worse performance than our main results in Table~\ref{tab:comparison-of-pretrained-and-ft-models}, where the representations from \emph{before} the dense layer are used. Using the representations from before the task-specific head also makes zero-shot OOD detection possible, where the task-specific head is randomly initialized, but weights from the backbone of the pre-trained model are used. 

\begin{table*}[ht!] 
\centering
\resizebox{\textwidth}{!}{ 
\begin{tabular}{llllllllll} 
\toprule
~ & ~ & \multicolumn{4}{c}{\textbf{KNN }(non-parametric)} & \multicolumn{4}{c}{\textbf{Mahalanobis}    
 (parametric)} \\ 

\textbf{ID$\to$OOD Pair} & \textbf{Training} & \textbf{AUROC ↑} & \textbf{AUPR (In) ↑} & \textbf{AUPR (Out) ↑} & \textbf{FPR95 ↓} & \textbf{AUROC ↑} & \textbf{AUPR (In) ↑} & \textbf{AUPR (Out) ↑} & \textbf{FPR95 ↓} \\ \midrule
\multicolumn{10}{l}{\textit{Out-of-Domain: Semantic Shift}} \\ \midrule
        
         & CE & 0.967 & 0.989 & 0.907 & 0.193 & 0.973 & 0.991 & 0.918 & 0.154 \\ 
        20NG$\to$SST-2 & TAPT & 0.962 & 0.988 & 0.885 & 0.226 & 0.971 & 0.990 & 0.911 & 0.164 \\ 
        ~ & SupCon & 0.962 & 0.987 & 0.889 & 0.230 & 0.971 & 0.990 & 0.917 & 0.159 \\ \midrule
         & CE & 0.946 & 0.884 & 0.981 & 0.311 & 0.955 & 0.900 & 0.984 & 0.250 \\ 
        20NG$\to$MNLI & TAPT & 0.942 & 0.875 & 0.980 & 0.314 & 0.952 & 0.887 & 0.983 & 0.253 \\ 
        ~ & SupCon & 0.946 & 0.884 & 0.981 & 0.311 & 0.957 & 0.904 & 0.985 & 0.246 \\ \midrule
         & CE & 0.912 & 0.953 & 0.839 & 0.445 & 0.927 & 0.960 & 0.870 & 0.373 \\ 
        20NG$\to$RTE & TAPT & 0.889 & 0.938 & 0.806 & 0.507 & 0.902 & 0.944 & 0.836 & 0.430 \\ 
        ~ & SupCon & 0.911 & 0.953 & 0.837 & 0.445 & 0.932 & 0.964 & 0.879 & 0.347 \\ \midrule
         & CE & 0.943 & 0.786 & 0.992 & 0.339 & 0.951 & 0.790 & 0.993 & 0.279 \\
        20NG$\to$IMDB & TAPT & 0.947 & 0.778 & 0.993 & 0.283 & 0.956 & 0.782 & 0.994 & 0.212 \\ 
        ~ & SupCon & 0.952 & 0.808 & 0.993 & 0.277 & 0.961 & 0.822 & 0.995 & 0.212 \\ \midrule
         & CE & 0.941 & 0.972 & 0.882 & 0.296 & 0.950 & 0.976 & 0.895 & 0.254 \\
        20NG$\to$Multi30K & TAPT & 0.932 & 0.967 & 0.870 & 0.313 & 0.942 & 0.971 & 0.891 & 0.247 \\
        ~ & SupCon & 0.928 & 0.964 & 0.869 & 0.331 & 0.940 & 0.970 & 0.892 & 0.274 \\ \midrule
         & CE & 0.932 & 0.864 & 0.974 & 0.375 & 0.941 & 0.878 & 0.978 & 0.324 \\
        20NG$\to$NewsCategory & TAPT & 0.924 & 0.844 & 0.971 & 0.384 & 0.933 & 0.852 & 0.975 & 0.326 \\
        ~ & SupCon & 0.929 & 0.861 & 0.973 & 0.396 & 0.944 & 0.886 & 0.979 & 0.319 \\ \midrule
         & CE & 0.946 & 0.990 & 0.783 & 0.285 & 0.952 & 0.991 & 0.800 & 0.255 \\ 
        20NG$\to$CLINC150 & TAPT & 0.935 & 0.987 & 0.739 & 0.343 & 0.945 & 0.989 & 0.774 & 0.280 \\
        ~ & SupCon & 0.932 & 0.987 & 0.732 & 0.372 & 0.943 & 0.989 & 0.770 & 0.319 \\
\midrule
\multicolumn{10}{l}{\textit{Out-of-Domain: Background Shift}} \\ \midrule
         & CE & 0.856 & 0.994 & 0.135 & 0.784 & 0.877 & 0.995 & 0.171 & 0.738 \\
        IMDB $\to$SST-2 & TAPT & 0.852 & 0.994 & 0.130 & 0.765 & 0.867 & 0.995 & 0.136 & 0.760 \\ 
        ~ & SupCon & 0.833 & 0.993 & 0.105 & 0.840 & 0.859 & 0.994 & 0.128 & 0.834 \\ 
\midrule
\multicolumn{10}{l}{\textit{Same Domain Shift}} \\ \midrule
        NewsCategory-ID $\to$ & CE & 0.924 & 0.924 & 0.930 & 0.499 & 0.887 & 0.837 & 0.914 & 0.490 \\ 
        NewsCategory-OOD & TAPT & 0.920 & 0.920 & 0.925 & 0.520 & 0.881 & 0.830 & 0.910 & 0.501 \\
        ~ & SupCon & 0.927 & 0.925 & 0.935 & 0.464 & 0.878 & 0.817 & 0.912 & 0.475 \\ 
\bottomrule
\end{tabular}}
\caption{Comparison of fine-tuning objectives with distance-based methods, using the representations from after the dense layer and before the classification projection layer.
}
\label{tab:comparison-of-ft-models-before-dense}
\end{table*}

%% file: sections/Appendix/Sentence-Embeddings.tex
\section{Generation of Sequence Embeddings}
\label{sec:appendix-sent-embeddings}

{Our experiments in the main paper use sentence embeddings obtained from the beginning-of-sentence (BOS) token. This practice is standard for most BERT-like models, including RoBERTa, which we use for our experiments. Prior work has also shown that using the average of all token embeddings can lead to the formation of similar domain-based clusters \cite{aharoni2020unsupervised}.}

{In this section, we compare this approach with the alternate approach of obtaining sequence embeddings as the average of all token embeddings in the sequence. Table~\ref{tab:sent-embedding-comparison} shows that both approaches yield almost identical performance on the OOD detection task.}

\begin{table}[t]
\centering
\resizebox{0.5\textwidth}{!}{
\begin{tabular}{llllll} 
\toprule
    \multicolumn{1}{c}{\textbf{OOD}} & \textbf{Embedding} & \textbf{AUROC (kNN) ↑}& \textbf{FPR (kNN) ↓} & \textbf{AUROC (kNN) ↑}& \textbf{FPR (kNN) ↓}\\
    \midrule
        SST-2 & Avg	& 1.000	& 1.000	& 1.000	& 0.000 \\ 
        ~   & BOS & 1.000 & 1.000 & 1.000 & 0.000 \\ \midrule
        MNLI & Avg & 1.000 & 0.999 & 1.000 & 0.000 \\
        ~   & BOS & 1.000 & 0.999 & 1.000 & 0.000 \\ \midrule
        RTE	 & Avg & 0.999 & 0.999 & 0.997 & 0.000 \\
        ~  & BOS & 1.000 & 1.000 & 0.999 & 0.000 \\ \midrule
        IMDB & Avg & 0.986 & 0.973 & 0.997 & 0.008 \\
        ~    & BOS & 0.988 & 0.970 & 0.998 & 0.019 \\ \midrule
        Multi30K & Avg & 1.000 & 1.000 & 1.000 & 0.000 \\ 
            ~    & BOS & 1.000 & 1.000 & 1.000 & 0.000 \\ \midrule
        NewsCategory & Avg & 1.000 & 0.999 & 1.000 & 0.000 \\ 
            ~        & BOS & 1.000 & 0.999 & 1.000 & 0.000 \\ \midrule
        CLINC150 & Avg & 1.000 & 1.000 & 1.000 & 0.000 \\ 
           ~     & BOS & 1.000 & 1.000 & 1.000 & 0.000 \\ \bottomrule
\end{tabular}}
\caption{{Comparison of methods to generate sequence embeddings. In the OoD Semantic Shift setting, where \texttt{20NewsGroups} is the ID dataset, the performance between Avg (averaging all token embeddings to get the sequence embedding) and BOS (using the first token embedding as the sequence embedding) are almost identical.}}
\label{tab:sent-embedding-comparison}
\vskip -0.2in
\end{table}

%% file: sections/Appendix/Detailed-Performance-of-Fine-tuning-for-OOD-Detection.tex
\section{Detailed Performance of Fine-tuning for OOD Detection}
\label{sec:appendix-finetuning-trajectories}

Table \ref{tab:ft-trajectories-20ng-to-rte} summarizes the epoch-wise performance when fine-tuning on ID data, for the setting of OoD semantic shift. Table \ref{tab:ft-trajectories-imdb-to-sst2} shows the same for OoD background shift, while Table \ref{tab:ft-trajectories-nc} shows this for same-domain (SD) shift.

\begin{table*}[ht]
\vskip 0.2in
\centering
\resizebox{\textwidth}{!}{
    \begin{tabular}{lllllllllllllll}
    \toprule
        \textbf{Training} & \textbf{Epoch} & \textbf{ID Accuracy ↑} & \textbf{Dispersion ↑} & \textbf{Compactness ↓} & \textbf{ID-OOD} & \multicolumn{2}{c}{\textbf{MSP}} & \multicolumn{2}{c}{\textbf{Energy}} & \multicolumn{2}{c}{\textbf{KNN}} & \multicolumn{2}{c}{\textbf{Mahalanobis}} \\ 
        ~ & ~ & ~ & ~ & ~ & \textbf{Separability ↑} & \textbf{AUROC ↑} & \textbf{FPR95 ↓} & \textbf{AUROC ↑} & \textbf{FPR95 ↓} & \textbf{AUROC ↑} & \textbf{FPR95 ↓} & \textbf{AUROC ↑} & \textbf{FPR95 ↓} \\ \midrule
         & 1 & 0.791 & 89.777 & 24.303 & 26.594 & 0.757 & 0.687 & 0.849 & 0.432 & 0.934 & 0.332 & 0.961 & 0.221 \\ 
        ~ & 2 & 0.823 & 90.632 & 22.508 & 26.595 & 0.790 & 0.656 & 0.855 & 0.421 & 0.925 & 0.373 & 0.956 & 0.247 \\ 
        ~ & 3 & 0.840 & 91.439 & 20.312 & 28.570 & 0.808 & 0.638 & 0.864 & 0.426 & 0.931 & 0.344 & 0.957 & 0.229 \\ 
        ~ & 4 & 0.851 & 91.934 & 18.293 & 29.259 & 0.816 & 0.658 & 0.859 & 0.432 & 0.931 & 0.356 & 0.958 & 0.238 \\ 
        CE & 5 & 0.843 & 91.643 & 17.757 & 29.247 & 0.808 & 0.672 & 0.854 & 0.450 & 0.928 & 0.367 & 0.953 & 0.243 \\ 
        ~ & 6 & 0.855 & 91.966 & 16.464 & 29.579 & 0.824 & 0.655 & 0.855 & 0.437 & 0.922 & 0.380 & 0.946 & 0.262 \\ 
        ~ & 7 & 0.856 & 92.097 & 16.210 & 29.064 & 0.832 & 0.691 & 0.862 & 0.459 & 0.919 & 0.422 & 0.942 & 0.277 \\ 
        ~ & 8 & 0.859 & 92.170 & 15.122 & 28.968 & 0.829 & 0.695 & 0.854 & 0.472 & 0.920 & 0.413 & 0.945 & 0.290 \\ 
        ~ & 9 & 0.858 & 92.211 & 14.745 & 30.084 & 0.841 & 0.653 & 0.863 & 0.448 & 0.925 & 0.393 & 0.946 & 0.274 \\ 
        ~ & 10 & 0.858 & 92.232 & 14.261 & 29.733 & 0.833 & 0.684 & 0.853 & 0.469 & 0.922 & 0.410 & 0.945 & 0.285 \\ 
        \midrule
         & 1 & 0.807 & 90.555 & 23.987 & 27.595 & 0.785 & 0.646 & 0.861 & 0.403 & 0.929 & 0.326 & 0.955 & 0.239 \\ 
        ~ & 2 & 0.840 & 91.058 & 21.600 & 27.174 & 0.784 & 0.662 & 0.852 & 0.418 & 0.916 & 0.351 & 0.942 & 0.264 \\ 
        ~ & 3 & 0.841 & 91.473 & 20.052 & 29.920 & 0.823 & 0.610 & 0.875 & 0.386 & 0.931 & 0.323 & 0.948 & 0.250 \\ 
        ~ & 4 & 0.842 & 91.517 & 18.602 & 27.894 & 0.798 & 0.677 & 0.845 & 0.456 & 0.910 & 0.379 & 0.932 & 0.293 \\ 
        TAPT & 5 & 0.851 & 91.766 & 17.315 & 27.091 & 0.814 & 0.680 & 0.849 & 0.473 & 0.909 & 0.395 & 0.928 & 0.313 \\ 
        ~ & 6 & 0.852 & 91.916 & 16.551 & 28.467 & 0.819 & 0.666 & 0.844 & 0.487 & 0.908 & 0.421 & 0.926 & 0.330 \\ 
        ~ & 7 & 0.857 & 92.016 & 15.881 & 25.505 & 0.803 & 0.712 & 0.824 & 0.541 & 0.893 & 0.486 & 0.913 & 0.393 \\ 
        ~ & 8 & 0.860 & 92.122 & 14.934 & 26.382 & 0.799 & 0.701 & 0.820 & 0.516 & 0.897 & 0.457 & 0.918 & 0.364 \\ 
        ~ & 9 & 0.856 & 92.149 & 14.602 & 26.829 & 0.808 & 0.691 & 0.828 & 0.508 & 0.897 & 0.463 & 0.918 & 0.360 \\ 
        ~ & 10 & 0.861 & 92.211 & 14.364 & 27.151 & 0.807 & 0.695 & 0.826 & 0.493 & 0.898 & 0.455 & 0.919 & 0.352 \\ 
        \midrule
         & 1 & 0.763 & 87.389 & 26.510 & 26.239 & 0.771 & 0.622 & 0.866 & 0.404 & 0.936 & 0.327 & 0.970 & 0.180 \\ 
        ~ & 2 & 0.820 & 89.348 & 23.556 & 27.233 & 0.771 & 0.661 & 0.851 & 0.438 & 0.935 & 0.333 & 0.967 & 0.206 \\ 
        ~ & 3 & 0.838 & 90.452 & 21.171 & 26.267 & 0.760 & 0.710 & 0.832 & 0.487 & 0.928 & 0.350 & 0.962 & 0.230 \\ 
        ~ & 4 & 0.842 & 90.874 & 20.170 & 28.124 & 0.796 & 0.660 & 0.859 & 0.410 & 0.927 & 0.343 & 0.960 & 0.206 \\ 
        SupCon & 5 & 0.851 & 91.295 & 18.608 & 28.033 & 0.815 & 0.649 & 0.865 & 0.412 & 0.921 & 0.382 & 0.954 & 0.272 \\ 
        ~ & 6 & 0.852 & 91.342 & 18.493 & 30.519 & 0.832 & 0.616 & 0.883 & 0.370 & 0.934 & 0.304 & 0.960 & 0.206 \\ 
        ~ & 7 & 0.855 & 91.736 & 17.224 & 28.144 & 0.818 & 0.711 & 0.863 & 0.448 & 0.922 & 0.375 & 0.954 & 0.248 \\ 
        ~ & 8 & 0.853 & 91.828 & 16.390 & 28.809 & 0.825 & 0.676 & 0.863 & 0.441 & 0.921 & 0.386 & 0.950 & 0.253 \\ 
        ~ & 9 & 0.857 & 91.977 & 15.999 & 28.812 & 0.832 & 0.666 & 0.869 & 0.452 & 0.922 & 0.390 & 0.952 & 0.247 \\ 
        ~ & 10 & 0.862 & 92.016 & 15.624 & 28.713 & 0.833 & 0.683 & 0.869 & 0.447 & 0.923 & 0.393 & 0.952 & 0.248 \\ 
        \bottomrule
\end{tabular}}
\caption{Effect of fine-tuning by various objectives on OOD detection performance.
 With \texttt{20NewsGroups} as ID and \texttt{RTE} as OOD, this ID-OOD pair exhibits a out-of-domain semantic shift.}
\label{tab:ft-trajectories-20ng-to-rte}
\end{table*}

\begin{table*}[ht!]
\vskip 0.2in
\centering
\resizebox{\textwidth}{!}{
    \begin{tabular}{lllllllllllllll}
    \toprule
        \textbf{Training} & \textbf{Epoch} & \textbf{ID Accuracy ↑} & \textbf{Dispersion ↑} & \textbf{Compactness ↓} & \textbf{ID-OOD} & \multicolumn{2}{c}{\textbf{MSP}} & \multicolumn{2}{c}{\textbf{Energy}} & \multicolumn{2}{c}{\textbf{KNN}} & \multicolumn{2}{c}{\textbf{Mahalanobis}} \\ 
        ~ & ~ & ~ & ~ & ~ & \textbf{Separability ↑} & \textbf{AUROC ↑} & \textbf{FPR95 ↓} & \textbf{AUROC ↑} & \textbf{FPR95 ↓} & \textbf{AUROC ↑} & \textbf{FPR95 ↓} & \textbf{AUROC ↑} & \textbf{FPR95 ↓} \\ \midrule
        ~ & 1 & 0.938 & 87.041 & 21.787 & 8.437 & 0.699 & 0.868 & 0.675 & 0.873 & 0.894 & 0.432 & 0.951 & 0.254 \\ 
        ~ & 2 & 0.937 & 81.117 & 20.439 & 5.936 & 0.677 & 0.894 & 0.676 & 0.921 & 0.896 & 0.429 & 0.947 & 0.295 \\ 
        ~ & 3 & 0.937 & 97.130 & 18.534 & 10.150 & 0.767 & 0.852 & 0.765 & 0.856 & 0.866 & 0.539 & 0.931 & 0.344 \\ 
        ~ & 4 & 0.938 & 99.677 & 16.615 & 11.517 & 0.735 & 0.841 & 0.746 & 0.839 & 0.865 & 0.613 & 0.901 & 0.490 \\ 
        CE & 5 & 0.927 & 114.249 & 15.839 & 11.704 & 0.719 & 0.881 & 0.734 & 0.882 & 0.850 & 0.625 & 0.896 & 0.478 \\ 
        ~ & 6 & 0.936 & 111.093 & 15.514 & 10.819 & 0.743 & 0.853 & 0.748 & 0.854 & 0.831 & 0.671 & 0.886 & 0.541 \\ 
        ~ & 7 & 0.938 & 122.309 & 14.283 & 14.760 & 0.745 & 0.829 & 0.752 & 0.826 & 0.860 & 0.679 & 0.889 & 0.571 \\ 
        ~ & 8 & 0.938 & 124.571 & 14.686 & 15.711 & 0.784 & 0.811 & 0.793 & 0.812 & 0.872 & 0.674 & 0.899 & 0.556 \\ 
        ~ & 9 & 0.941 & 130.242 & 13.908 & 16.455 & 0.787 & 0.805 & 0.798 & 0.806 & 0.872 & 0.713 & 0.898 & 0.596 \\ 
        ~ & 10 & 0.939 & 130.285 & 14.314 & 15.770 & 0.781 & 0.813 & 0.794 & 0.813 & 0.865 & 0.741 & 0.893 & 0.618 \\ 
        \midrule
        ~ & 1 & 0.940 & 76.871 & 15.894 & 7.455 & 0.733 & 0.830 & 0.708 & 0.838 & 0.902 & 0.414 & 0.966 & 0.166 \\ 
        ~ & 2 & 0.943 & 82.230 & 15.106 & 10.080 & 0.805 & 0.808 & 0.803 & 0.820 & 0.918 & 0.418 & 0.960 & 0.242 \\ 
        ~ & 3 & 0.937 & 89.350 & 14.646 & 10.831 & 0.814 & 0.782 & 0.810 & 0.789 & 0.867 & 0.650 & 0.916 & 0.513 \\ 
        ~ & 4 & 0.938 & 100.884 & 13.629 & 11.705 & 0.810 & 0.792 & 0.802 & 0.795 & 0.866 & 0.644 & 0.898 & 0.583 \\ 
        TAPT & 5 & 0.940 & 116.726 & 12.179 & 12.610 & 0.790 & 0.820 & 0.781 & 0.820 & 0.863 & 0.679 & 0.887 & 0.595 \\ 
        ~ & 6 & 0.940 & 117.262 & 11.048 & 11.496 & 0.770 & 0.829 & 0.773 & 0.831 & 0.861 & 0.641 & 0.890 & 0.533 \\ 
        ~ & 7 & 0.940 & 119.857 & 10.796 & 13.009 & 0.789 & 0.806 & 0.789 & 0.810 & 0.870 & 0.634 & 0.901 & 0.519 \\ 
        ~ & 8 & 0.942 & 127.375 & 10.332 & 14.030 & 0.808 & 0.799 & 0.811 & 0.797 & 0.859 & 0.680 & 0.875 & 0.613 \\ 
        ~ & 9 & 0.944 & 134.293 & 8.886 & 14.992 & 0.787 & 0.792 & 0.791 & 0.790 & 0.859 & 0.738 & 0.881 & 0.682 \\ 
        ~ & 10 & 0.943 & 134.601 & 9.060 & 15.340 & 0.797 & 0.794 & 0.801 & 0.795 & 0.857 & 0.746 & 0.877 & 0.683 \\
        \midrule
        ~ & 1 & 0.928 & 135.550 & 19.245 & 11.282 & 0.669 & 0.869 & 0.667 & 0.876 & 0.855 & 0.600 & 0.930 & 0.381 \\ 
        ~ & 2 & 0.927 & 133.438 & 18.591 & 10.494 & 0.682 & 0.865 & 0.674 & 0.891 & 0.809 & 0.592 & 0.903 & 0.423 \\ 
        ~ & 3 & 0.929 & 148.985 & 13.544 & 9.218 & 0.708 & 0.872 & 0.698 & 0.882 & 0.807 & 0.696 & 0.876 & 0.621 \\ 
        ~ & 4 & 0.937 & 158.041 & 8.588 & 12.908 & 0.742 & 0.842 & 0.736 & 0.842 & 0.846 & 0.726 & 0.884 & 0.666 \\ 
        SupCon & 5 & 0.935 & 161.662 & 7.455 & 13.168 & 0.711 & 0.854 & 0.725 & 0.853 & 0.849 & 0.711 & 0.876 & 0.639 \\ 
        ~ & 6 & 0.937 & 163.736 & 6.264 & 11.734 & 0.752 & 0.865 & 0.732 & 0.865 & 0.849 & 0.742 & 0.877 & 0.698 \\ 
        ~ & 7 & 0.936 & 164.397 & 5.306 & 9.679 & 0.688 & 0.868 & 0.678 & 0.868 & 0.849 & 0.775 & 0.877 & 0.744 \\ 
        ~ & 8 & 0.938 & 167.184 & 4.434 & 9.826 & 0.749 & 0.850 & 0.726 & 0.852 & 0.842 & 0.793 & 0.870 & 0.774 \\ 
        ~ & 9 & 0.938 & 167.316 & 4.306 & 8.397 & 0.727 & 0.858 & 0.745 & 0.859 & 0.841 & 0.815 & 0.868 & 0.787 \\ 
        ~ & 10 & 0.938 & 167.586 & 4.182 & 8.259 & 0.720 & 0.851 & 0.736 & 0.851 & 0.838 & 0.824 & 0.865 & 0.800 \\ \bottomrule
\end{tabular}}
\caption{Effect of fine-tuning by various objectives on OOD detection performance. With \texttt{IMDB} as ID and \texttt{SST-2} as OOD, this ID-OOD pair exhibits a out-of-domain background shift.}
\label{tab:ft-trajectories-imdb-to-sst2}
\end{table*}

\begin{table*}
\vskip 0.2in
\centering
\resizebox{\textwidth}{!}{
    \begin{tabular}{lllllllllllllll}
    \toprule
        \textbf{Training} & \textbf{Epoch} & \textbf{ID Accuracy ↑} & \textbf{Dispersion ↑} & \textbf{Compactness ↓} & \textbf{ID-OOD} & \multicolumn{2}{c}{\textbf{MSP}} & \multicolumn{2}{c}{\textbf{Energy}} & \multicolumn{2}{c}{\textbf{KNN}} & \multicolumn{2}{c}{\textbf{Mahalanobis}} \\ 
        ~ & ~ & ~ & ~ & ~ & \textbf{ Separability ↑} & \textbf{AUROC ↑} & \textbf{FPR95 ↓} & \textbf{AUROC ↑} & \textbf{FPR95 ↓} & \textbf{AUROC ↑} & \textbf{FPR95 ↓} & \textbf{AUROC ↑} & \textbf{FPR95 ↓} \\ \midrule
         & 1 & 0.745 & 86.386 & 38.342 & 13.311 & 0.739 & 0.794 & 0.810 & 0.705 & 0.927 & 0.481 & 0.829 & 0.626 \\ 
        ~ & 2 & 0.804 & 87.198 & 35.562 & 14.676 & 0.733 & 0.787 & 0.810 & 0.692 & 0.929 & 0.475 & 0.847 & 0.609 \\ 
        ~ & 3 & 0.842 & 89.052 & 33.008 & 17.263 & 0.749 & 0.770 & 0.819 & 0.636 & 0.934 & 0.446 & 0.867 & 0.547 \\ 
        ~ & 4 & 0.860 & 89.508 & 30.364 & 18.668 & 0.750 & 0.780 & 0.822 & 0.629 & 0.933 & 0.446 & 0.878 & 0.520 \\ 
        CE & 5 & 0.872 & 91.260 & 29.191 & 18.844 & 0.794 & 0.752 & 0.842 & 0.603 & 0.927 & 0.473 & 0.872 & 0.525 \\ 
        ~ & 6 & 0.878 & 90.918 & 27.667 & 19.017 & 0.798 & 0.736 & 0.834 & 0.607 & 0.921 & 0.495 & 0.865 & 0.515 \\ 
        ~ & 7 & 0.884 & 91.440 & 25.515 & 21.154 & 0.821 & 0.706 & 0.855 & 0.549 & 0.927 & 0.469 & 0.885 & 0.475 \\ 
        ~ & 8 & 0.888 & 91.601 & 24.952 & 21.588 & 0.830 & 0.700 & 0.858 & 0.555 & 0.925 & 0.500 & 0.885 & 0.475 \\ 
        ~ & 9 & 0.890 & 91.885 & 24.063 & 21.728 & 0.837 & 0.693 & 0.862 & 0.548 & 0.924 & 0.499 & 0.884 & 0.474 \\ 
        ~ & 10 & 0.890 & 91.969 & 23.580 & 22.184 & 0.844 & 0.676 & 0.866 & 0.541 & 0.924 & 0.489 & 0.887 & 0.479 \\  \midrule
         & 1 & 0.756 & 85.080 & 38.572 & 13.219 & 0.737 & 0.800 & 0.794 & 0.750 & 0.924 & 0.500 & 0.832 & 0.631 \\ 
        ~ & 2 & 0.825 & 87.712 & 35.636 & 15.552 & 0.734 & 0.782 & 0.811 & 0.678 & 0.928 & 0.493 & 0.854 & 0.587 \\ 
        ~ & 3 & 0.852 & 89.502 & 33.618 & 18.240 & 0.780 & 0.728 & 0.835 & 0.609 & 0.933 & 0.438 & 0.874 & 0.508 \\ 
        ~ & 4 & 0.874 & 89.802 & 31.870 & 18.473 & 0.777 & 0.754 & 0.828 & 0.601 & 0.926 & 0.463 & 0.869 & 0.523 \\ 
        TAPT & 5 & 0.886 & 91.409 & 29.624 & 18.564 & 0.792 & 0.737 & 0.830 & 0.830 & 0.917 & 0.518 & 0.855 & 0.573 \\ 
        ~ & 6 & 0.882 & 91.537 & 28.103 & 19.632 & 0.812 & 0.723 & 0.841 & 0.587 & 0.918 & 0.523 & 0.863 & 0.531 \\ 
        ~ & 7 & 0.891 & 91.683 & 26.551 & 20.700 & 0.823 & 0.711 & 0.853 & 0.559 & 0.924 & 0.486 & 0.875 & 0.503 \\ 
        ~ & 8 & 0.889 & 91.731 & 25.830 & 20.536 & 0.829 & 0.694 & 0.851 & 0.574 & 0.918 & 0.515 & 0.869 & 0.524 \\ 
        ~ & 9 & 0.888 & 91.874 & 25.309 & 21.490 & 0.835 & 0.683 & 0.858 & 0.563 & 0.920 & 0.494 & 0.878 & 0.489 \\ 
        ~ & 10 & 0.890 & 91.969 & 24.302 & 21.409 & 0.839 & 0.686 & 0.858 & 0.556 & 0.918 & 0.513 & 0.875 & 0.502 \\
        \midrule
         & 1 & 0.667 & 69.588 & 36.713 & 9.288 & 0.734 & 0.796 & 0.786 & 0.726 & 0.922 & 0.510 & 0.820 & 0.656 \\ 
        ~ & 2 & 0.750 & 75.252 & 34.277 & 11.627 & 0.748 & 0.742 & 0.808 & 0.669 & 0.926 & 0.496 & 0.827 & 0.619 \\ 
        ~ & 3 & 0.803 & 79.054 & 31.839 & 13.914 & 0.738 & 0.771 & 0.806 & 0.674 & 0.935 & 0.437 & 0.856 & 0.561 \\ 
        ~ & 4 & 0.822 & 82.853 & 29.858 & 15.612 & 0.741 & 0.769 & 0.807 & 0.652 & 0.931 & 0.445 & 0.856 & 0.555 \\ 
        SupCon & 5 & 0.847 & 84.920 & 28.296 & 17.149 & 0.748 & 0.774 & 0.803 & 0.638 & 0.929 & 0.452 & 0.863 & 0.520 \\ 
        ~ & 6 & 0.868 & 88.327 & 26.281 & 18.311 & 0.774 & 0.757 & 0.808 & 0.637 & 0.923 & 0.470 & 0.863 & 0.524 \\ 
        ~ & 7 & 0.869 & 89.118 & 24.956 & 19.524 & 0.790 & 0.747 & 0.823 & 0.587 & 0.926 & 0.462 & 0.872 & 0.500 \\ 
        ~ & 8 & 0.882 & 89.527 & 24.449 & 20.277 & 0.794 & 0.722 & 0.827 & 0.584 & 0.927 & 0.449 & 0.874 & 0.471 \\ 
        ~ & 9 & 0.884 & 90.408 & 23.481 & 20.775 & 0.813 & 0.711 & 0.836 & 0.581 & 0.924 & 0.473 & 0.873 & 0.467 \\ 
        ~ & 10 & 0.884 & 90.487 & 23.106 & 21.220 & 0.821 & 0.697 & 0.842 & 0.568 & 0.925 & 0.465 & 0.877 & 0.465 \\\midrule
\end{tabular}}
\vspace{-2mm}
\caption{Effect of fine-tuning by various objectives on OOD detection performance. Using subsets of the \texttt{NewsCategory} as ID and OOD, this ID-OOD pair exhibits a same-domain shift.}
\vskip -0.2in
\label{tab:ft-trajectories-nc}
\end{table*}

%% file: sections/Appendix/Effect-of-Temperature-in-SupCon.tex
\section{Effect of Temperature in SupCon}
\label{sec:appendix-supcon-temperature-study}

\begin{figure}[ht]
\vskip 0.2in
\begin{center}
\centerline{
\includegraphics[width=0.5\columnwidth]{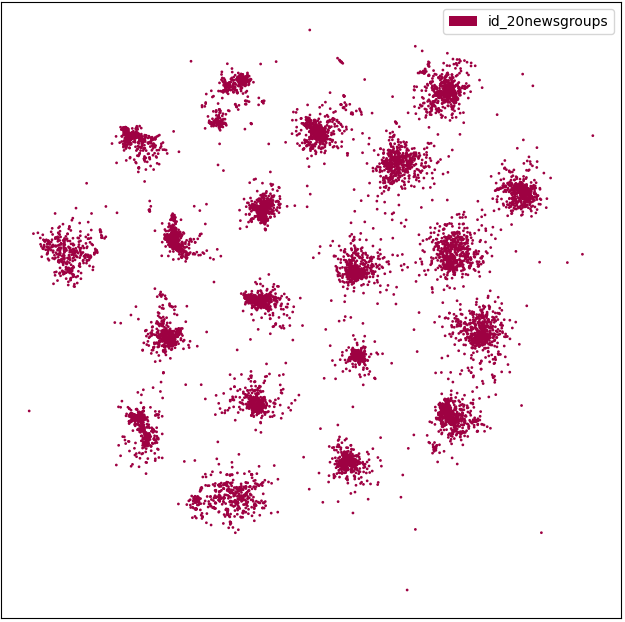}
\includegraphics[width=0.5\columnwidth]{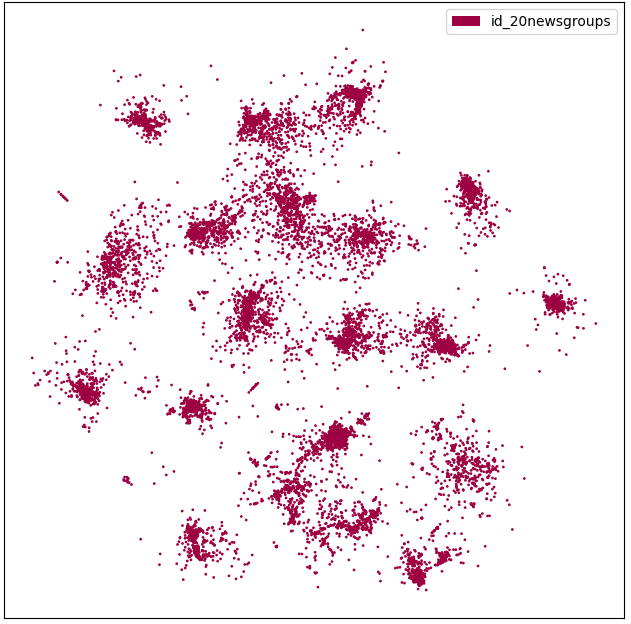}
}
\caption{Effect of the temperature $\tau$ on representations  trained with the SupCon loss. The ID data is \texttt{20NewsGroups}. \textbf{Left}: $\tau=0.1$. \textbf{Right}: $\tau=0.7$.}
\label{fig:20ng-viz-effect-of-tau}
\end{center}
\vskip -0.35in
\end{figure}

Contrastive loss is shown to be a hardness-aware loss function, penalizing hard negative samples by reducing tolerance to them \cite{wang2021understanding}. The temperature $\tau$ has been shown to control the tolerance to negative samples. As seen in Figure \ref{fig:20ng-viz-effect-of-tau}, low temperature leads to a uniform distribution with high separability in the learnt embedding space, but this can reduce tolerance to semantically similar samples, breaking underlying semantic structure. The temperature must be set optimally to balance the `uniformity-tolerance' trade-off, having some tolerance to semantically similar examples. 
When \texttt{IMDB} is ID, we find OOD detection to be optimal at $\tau=0.7$, since the two classes of the dataset share semantic similarities. However, with the \texttt{20NewsGroups} topic classification task, we find a lower value of $\tau=0.1$ to be optimal. This is because a larger number of ID classes requires a stronger uniformity in the learnt distribution, and the weaker semantic similarities between classes assures that this uniformity does not hurt performance. 

Tables \ref{tab:tau-sweep-20ng-rte}, 
\ref{tab:tau-sweep-nc} and \ref{tab:tau-sweep-imdb-sst2} show the effects of varying the temperature parameter $\tau$ in the SupCon loss, on OOD detection, in the settings of OoD semantic shift, OoD background shift and same-domain shift. All models are fine-tuned for 10 epochs. 

%% file: sections/Appendix/Effect-of-k.tex
\section{Effect of \texorpdfstring{$k$}{k}}
\label{sec:appendix-knn-analysis}

\begin{figure*}[htb!]
\vskip 0.2in
\begin{center}
\centerline{
\includegraphics[width=0.5\textwidth]{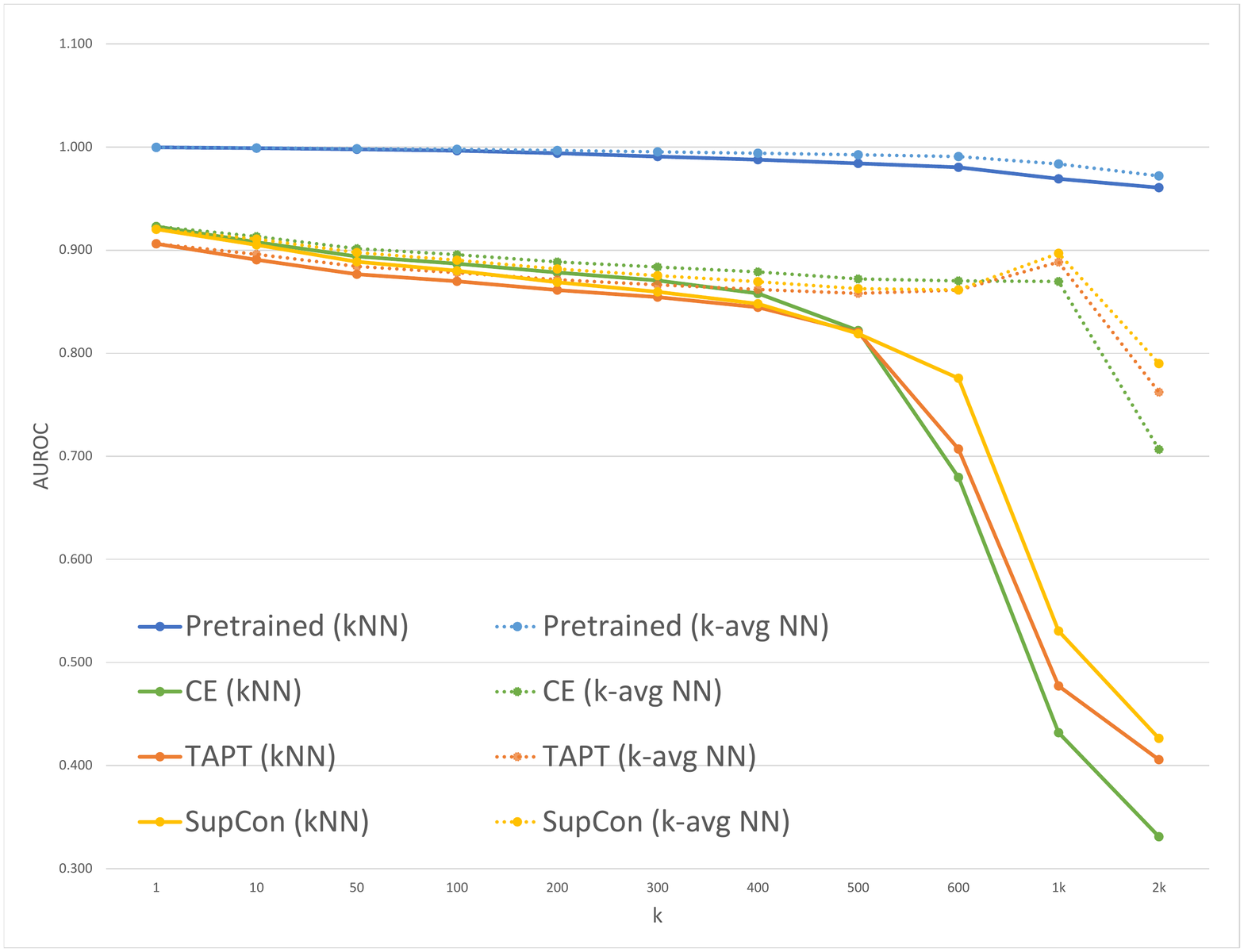}
\includegraphics[width=0.5\textwidth]{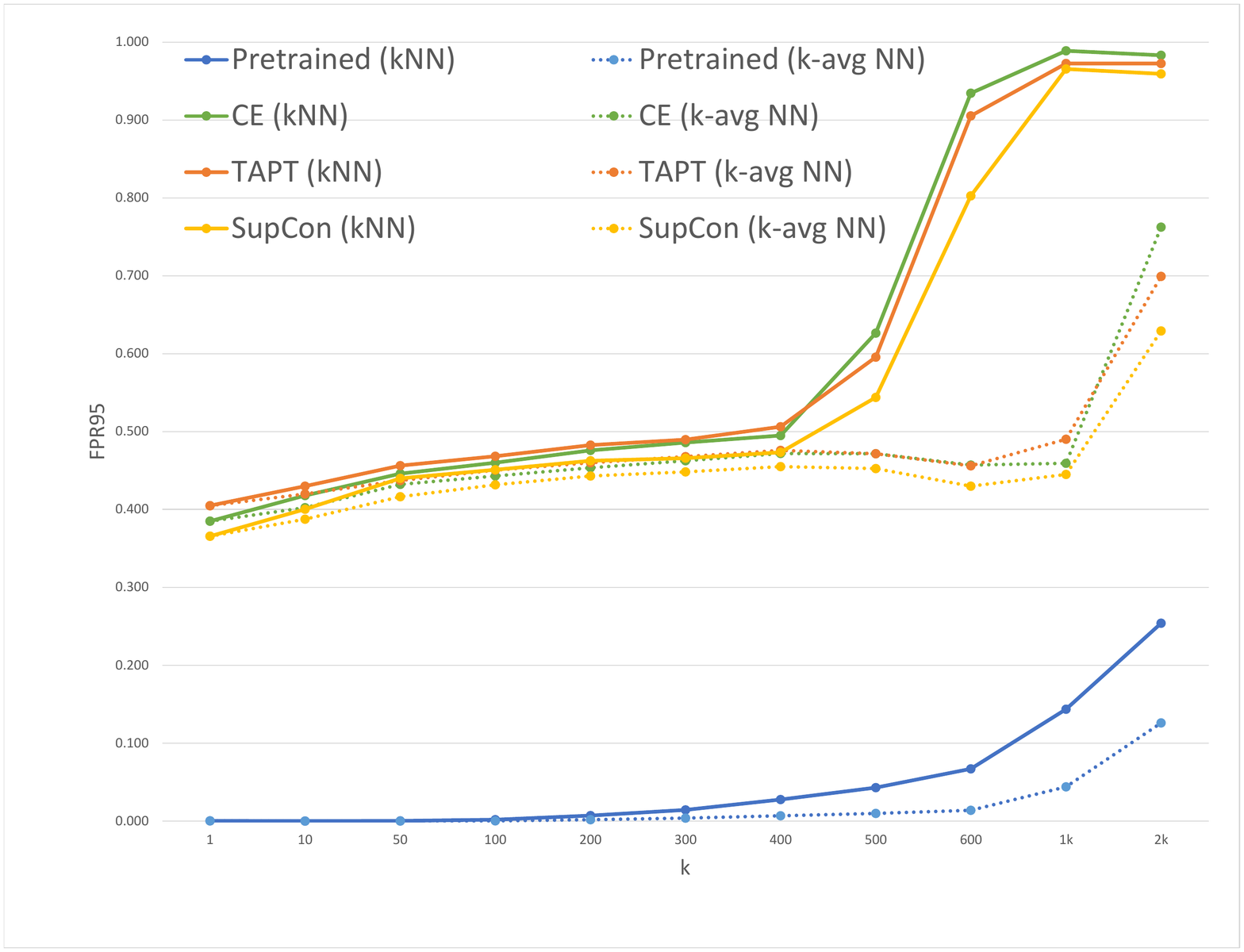}
}
\caption{Effect of $k$ in OOD detection using kNN, for the OoD semantic shift setting (\texttt{20NewsGroups}$\to$\texttt{RTE}). Left: AUROC. Right: FPR95.}
\label{fig:knn-sweep}
\end{center}
\vskip -0.35in
\end{figure*}

Figure \ref{fig:knn-sweep} shows us that $k=1$ is consistently the optimal $k$ for kNN, across fine-tuning objectives and distribution shifts. The detection performance remains strong until $k$ reaches the ID class size, which is between 400 and 600 for \texttt{20NewsGroups}. After this point, the nearest neighbour for an ID and OOD point will both be outside the nearest ID class cluster, making both distances more comparable and harder to distinguish. With pre-trained models, the performance remains strong as there is no concept of class clusters and a single domain cluster is instead present.

%% file: tables/implementation-details.tex
\begin{table*}[t!]
\vskip 0.2in
\centering
\resizebox{\textwidth}{!}{
\begin{tabular}{llllllllll} 
\toprule
~ & ~ & \multicolumn{4}{c}{\textbf{KNN}(non-parametric)} & \multicolumn{4}{c}{\textbf{Mahalanobis}    
 (parametric)} \\

\textbf{ID$\to$OOD Pair} & \textbf{Training} & \textbf{AUROC ↑} & \textbf{AUPR (In) ↑} & \textbf{AUPR (Out) ↑} & \textbf{FPR95 ↓} & \textbf{AUROC ↑} & \textbf{AUPR (In) ↑} & \textbf{AUPR (Out) ↑} & \textbf{FPR95 ↓} \\ \midrule
\multicolumn{10}{l}{\textit{Out-of-Domain: Semantic Shift}} \\ \midrule
        20NG$\to$SST-2 & CE & 0.973 & 0.991 & 0.923 & 0.155 & 0.981 & 0.994 & 0.942 & 0.087 \\ 
        ~ & TAPT & 0.969 & 0.990 & 0.903 & 0.169 & 0.981 & 0.994 & 0.939 & 0.088 \\
        ~ & SupCon & 0.969 & 0.990 & 0.909 & 0.180 & 0.980 & 0.994 & 0.943 & 0.094 \\ 
        \rowcolor{COLOR_ZS} \cellcolor{white} ~ & Pre-trained & 1.000 & 1.000 & 1.000 & 0.000 & 1.000 & 1.000 & 1.000 & 0.000 \\ \midrule
        
        20NG$\to$RTE & CE & 0.922 & 0.958 & 0.858 & 0.410 & 0.945 & 0.970 & 0.902 & 0.285  \\
        ~ & TAPT & 0.898 & 0.942 & 0.822 & 0.455 & 0.919 & 0.952 & 0.869 & 0.352 \\ 
        ~ & SupCon & 0.923 & 0.959 & 0.858 & 0.393 & 0.952 & 0.975 & 0.914 & 0.248 \\ 
         \rowcolor{COLOR_ZS} \cellcolor{white} ~ & Pre-trained  & 1.000 & 1.000 & 0.999 & 0.000 & 1.000 & 1.000 & 0.999 & 0.000 \\ 
\bottomrule
\end{tabular}}
\caption{Comparison of OOD detection performance of pre-trained and fine-tuned models, averaged over 3 runs.}
\label{tab:average-over-3-runs}
\end{table*}

\begin{table}
\centering
\resizebox{\columnwidth}{!}{
    \begin{tabular}{ll}
    \toprule
        Hyperparameter & Value \\ \midrule
        Batch size & 4 \\
        Learning rate & 1e-5 \\
        Weight decay & 0.01 \\
        Maximum sequence length & 256 \\
        Number of pre-training epochs (for TAPT) & 3 \\
        Contrastive loss weight (for SupCon) & 2.0 \\
        CE loss weight (for SupCon) & 1.0 \\
        Temperature (for SupCon) & 0.1 or 0.7 ($*$) \\
        \bottomrule
    \end{tabular}
    }
\caption{Hyperparameters used in our study. ($*$) Values depend on the dataset.}
\label{tab:hyperparameters}
\end{table}

%% file: sections/Appendix/Details-on-Implementation.tex
\section{Details on Implementation}
\label{sec:appendix-checklist-implementation}

We use RoBERTa from the HuggingFace library\footnote{\url{https://github.com/huggingface/transformers}}, and use PyTorch to train our models. Hyperparameter search is performed through a grid search. Apart from the default parameters in the trainer module from HuggingFace, our selected hyperparameters are listed in Table~\ref{tab:hyperparameters}.

\vspace{-0.3cm}
\paragraph{Computations} The RoBERTa base model has approximately 125 million parameters, including those of the classification head. On a single NVIDIA GeForce RTX 2080 Ti GPU, training the model for 10 epochs takes approximately 8-12 hours, and OOD detection for a single dataset takes approximately 15 minutes. Over the scale of our experiments, we have used about 200 hours of GPU training time.

\vspace{-0.3cm}
\paragraph{Multiple Runs} Following the protocol in~\citet{arora2021types}, we report results over a single run. However, in Table~\ref{tab:average-over-3-runs} we show results of a subset of experiments averaged over 3 runs. There is no significant difference between the results in Table~\ref{tab:average-over-3-runs} and Table~\ref{tab:comparison-of-pretrained-and-ft-models}, indicating that our experiments are stable across runs. Therefore, for the sake of computational resources and time, we stick to the single-run practice in our experiments.
\vspace{0.15in}

%% file: tables/supcon-temperature-effect.tex
\begin{table}
\centering
\resizebox{0.5\textwidth}{!}{
\begin{tabular}{llllllllll}
\toprule
\textbf{$\tau$} & \textbf{ID Acc.} & \multicolumn{2}{c}{\textbf{MSP}} & \multicolumn{2}{c}{\textbf{Energy}} & \multicolumn{2}{c}{\textbf{KNN}} & \multicolumn{2}{c}{\textbf{Mahalanobis}} \\ 
~ & ~ & \textbf{AUROC↑} & \textbf{FPR95↓} & \textbf{AUROC↑} & \textbf{FPR95↓} & \textbf{AUROC↑} & {FPR95} & \textbf{AUROC↑} & {FPR95↓} \\ \midrule
        0.1 & 0.851 & 0.830 & 0.662 & 0.868 & 0.413 & 0.913 & 0.413 & 0.930 & 0.349 \\
        0.2 & 0.850 & 0.826 & 0.635 & 0.851 & 0.422 & 0.910 & 0.426 & 0.932 & 0.316 \\
        0.3 & 0.855 & 0.839 & 0.650 & 0.864 & 0.447 & 0.913 & 0.448 & 0.933 & 0.342 \\
        0.4 & 0.853 & 0.817 & 0.671 & 0.836 & 0.486 & 0.905 & 0.470 & 0.925 & 0.373 \\
        0.5 & 0.853 & 0.822 & 0.645 & 0.844 & 0.441 & 0.904 & 0.434 & 0.921 & 0.347 \\
        0.6 & 0.852 & 0.816 & 0.649 & 0.836 & 0.475 & 0.901 & 0.453 & 0.918 & 0.364 \\
        0.7 & 0.853 & 0.805 & 0.683 & 0.822 & 0.518 & 0.887 & 0.495 & 0.903 & 0.417 \\
        0.8 & 0.854 & 0.805 & 0.673 & 0.827 & 0.506 & 0.903 & 0.468 & 0.920 & 0.394 \\
        0.9 & 0.854 & 0.818 & 0.668 & 0.840 & 0.483 & 0.902 & 0.483 & 0.920 & 0.399 \\
        1 & 0.853 & 0.799 & 0.706 & 0.814 & 0.509 & 0.894 & 0.489 & 0.912 & 0.400 \\ \bottomrule
\end{tabular}
}
\caption{Effect of the temperature $\tau$ in SupCon fine-tuning, on OOD detection, for OoD semantic shift  (\texttt{20NewsGroups}$\to$\texttt{RTE}).}
\label{tab:tau-sweep-20ng-rte}
\end{table}

\begin{table}[h]
\centering
\resizebox{0.5\textwidth}{!}{
\begin{tabular}{llllllllll}
\toprule
\textbf{$\tau$} & \textbf{ID Acc.} & \multicolumn{2}{c}{\textbf{MSP}} & \multicolumn{2}{c}{\textbf{Energy}} & \multicolumn{2}{c}{\textbf{KNN}} & \multicolumn{2}{c}{\textbf{Mahalanobis}} \\ 
~ & ~ & \textbf{AUROC↑} & \textbf{FPR95↓} & \textbf{AUROC↑} & \textbf{FPR95↓} & \textbf{AUROC↑} & {FPR95} & \textbf{AUROC↑} & {FPR95↓} \\ \midrule
        0.1 & 0.939 & 0.788 & 0.833 & 0.728 & 0.836 & 0.842 & 0.750 & 0.866 & 0.750 \\ 
        0.2 & 0.940 & 0.682 & 0.850 & 0.642 & 0.852 & 0.819 & 0.812 & 0.844 & 0.796 \\ 
        0.3 & 0.941 & 0.725 & 0.835 & 0.732 & 0.834 & 0.832 & 0.814 & 0.856 & 0.792 \\ 
        0.4 & 0.939 & 0.751 & 0.859 & 0.721 & 0.861 & 0.822 & 0.835 & 0.845 & 0.812 \\ 
        0.5 & 0.940 & 0.784 & 0.842 & 0.758 & 0.837 & 0.826 & 0.825 & 0.849 & 0.796 \\ 
        0.6 & 0.939 & 0.768 & 0.818 & 0.719 & 0.820 & 0.829 & 0.797 & 0.855 & 0.776 \\ 
        0.7 & 0.938 & 0.720 & 0.851 & 0.736 & 0.851 & 0.833 & 0.833 & 0.859 & 0.834 \\ 
        0.8 & 0.940 & 0.775 & 0.828 & 0.651 & 0.826 & 0.823 & 0.820 & 0.841 & 0.806 \\ 
        0.9 & 0.939 & 0.757 & 0.891 & 0.652 & 0.889 & 0.861 & 0.829 & 0.876 & 0.811 \\ 
        1 & 0.939 & 0.738 & 0.857 & 0.748 & 0.857 & 0.809 & 0.835 & 0.840 & 0.822 \\ \bottomrule
\end{tabular}
}
\caption{Effect of the temperature $\tau$ in SupCon fine-tuning, on OOD detection, for OoD background shift (\texttt{IMDB}$\rightarrow$\texttt{SST-2}).}
\label{tab:tau-sweep-imdb-sst2}
\end{table}

\begin{table}[b!]
\centering
\resizebox{0.5\textwidth}{!}{
\begin{tabular}{llllllllll}
\hline
\textbf{$\tau$} & \textbf{ID Acc.} & \multicolumn{2}{c}{\textbf{MSP}} & \multicolumn{2}{c}{\textbf{Energy}} & \multicolumn{2}{c}{\textbf{KNN}} & \multicolumn{2}{c}{\textbf{Mahalanobis}} \\ 
~ & ~ & \textbf{AUROC↑} & \textbf{FPR95↓} & \textbf{AUROC↑} & \textbf{FPR95↓} & \textbf{AUROC↑} & {FPR95} & \textbf{AUROC↑} & {FPR95↓} \\ \midrule
        0.1 & 0.888 & 0.817 & 0.700 & 0.842 & 0.570 & 0.927 & 0.470 & 0.877 & 0.478 \\ 
        0.2 & 0.885 & 0.825 & 0.681 & 0.835 & 0.592 & 0.922 & 0.509 & 0.878 & 0.510 \\ 
        0.3 & 0.879 & 0.802 & 0.733 & 0.817 & 0.600 & 0.922 & 0.502 & 0.866 & 0.525 \\ 
        0.4 & 0.889 & 0.815 & 0.670 & 0.809 & 0.594 & 0.922 & 0.522 & 0.874 & 0.524 \\ 
        0.5 & 0.822 & 0.706 & 0.818 & 0.749 & 0.747 & 0.913 & 0.576 & 0.821 & 0.662 \\ 
        0.6 & 0.890 & 0.794 & 0.713 & 0.796 & 0.641 & 0.919 & 0.561 & 0.871 & 0.563 \\ 
        0.7 & 0.891 & 0.811 & 0.694 & 0.804 & 0.609 & 0.921 & 0.534 & 0.876 & 0.538 \\ 
        0.8 & 0.892 & 0.814 & 0.697 & 0.812 & 0.602 & 0.922 & 0.534 & 0.879 & 0.525 \\ 
        0.9 & 0.847 & 0.730 & 0.798 & 0.747 & 0.714 & 0.909 & 0.606 & 0.818 & 0.677 \\ 
        1 & 0.888 & 0.817 & 0.706 & 0.819 & 0.611 & 0.920 & 0.534 & 0.875 & 0.541 \\ 
        \bottomrule
\end{tabular}
}
\caption{Effect of the temperature $\tau$ in SupCon fine-tuning, on OOD detection, for same-domain shift with the \texttt{NewsCategory} dataset.}
\label{tab:tau-sweep-nc}
\end{table}